\documentclass[10pt,twocolumn,letterpaper]{article}

\usepackage{icb}
\usepackage{times}
\usepackage{epsfig}
\usepackage{amsmath,graphicx}
\usepackage{float}
\newcommand{\red}[1]{\color{red}{#1}}
\newcommand{\blu}[1]{\color{blue}{#1}}

\usepackage{multirow}
\usepackage{enumitem} 
\usepackage{rotating}
\usepackage{makecell}
\usepackage[english]{babel}
\usepackage{float}

\icbfinalcopy 

\ificbfinal\fi
\begin{document}

\title{Face Hallucination Revisited: An Exploratory Study on Dataset Bias}

%
%
\author{Klemen grm$^\dagger$, Martin Pernu\v{s}$^\dagger$, Leo Cluzel$^\ddag$, Walter Scheirer$^\ast$, Simon Dobri\v{s}ek$^\dagger$, Vitomir \v{S}truc$^\dagger$\\
$^\dagger$University of Ljubljana, $^\ddag$ENSEA, $^\ast$University of Notre Dame\\
{\tt\small klemen.grm@fe.uni-lj.si}}

\maketitle
\thispagestyle{empty}


\begin{abstract}
    Contemporary face hallucination (FH) models exhibit considerable ability to reconstruct high-resolution (HR) details from low-resolution (LR) face images.  This ability is commonly learned from examples of corresponding HR-LR image pairs, created by artificially down-sampling the HR ground truth data. This down-sampling (or degradation) procedure not only defines the characteristics of the LR training data, but also determines the type of image degradations the learned FH models are eventually able to handle.
    If the image characteristics encountered with real-world LR images differ from the ones seen during training, FH models are still expected to perform well, but in practice may not produce the desired results. In this paper we study this problem and explore the bias introduced into FH models by the characteristics of the training data. We systematically analyze the generalization capabilities of several FH models in various scenarios, where the image the degradation function does not match the training setup and conduct experiments with synthetically downgraded as well as real-life low-quality images. We make several interesting findings that provide insight into existing problems with FH models and point to future research directions. 
\end{abstract}

\section{Introduction}
\label{sec:intro}
Face hallucination 
(FH) refers to the task of recovering high-resolution (HR) facial images from corresponding low-resolution (LR) inputs~\cite{baker_pami2002, chen2018fsrnet, grm2018face}. Solutions to this task have applications in face-oriented vision problems, such as face editing and alignment, 3D reconstruction or face  attribute estimation~\cite{Super-FAN,chen2018fsrnet, jourabloo2017pose, li2017generative, lin2007super,QLC,roth2016adaptive, yu2018super} and are used to mitigate performance degradations caused by input images of insufficient resolution. One particularly popular use of FH models is for LR face recognition tasks\cite{gunturk2003eigenface,lin2007super, zhao2015heterogeneous}, where LR probe images are super-resolved to reduce the dissimilarity with HR gallery data.
\begin{figure}[t!]
\centering
\begin{minipage}{0.96\columnwidth}
\includegraphics[width=\textwidth]{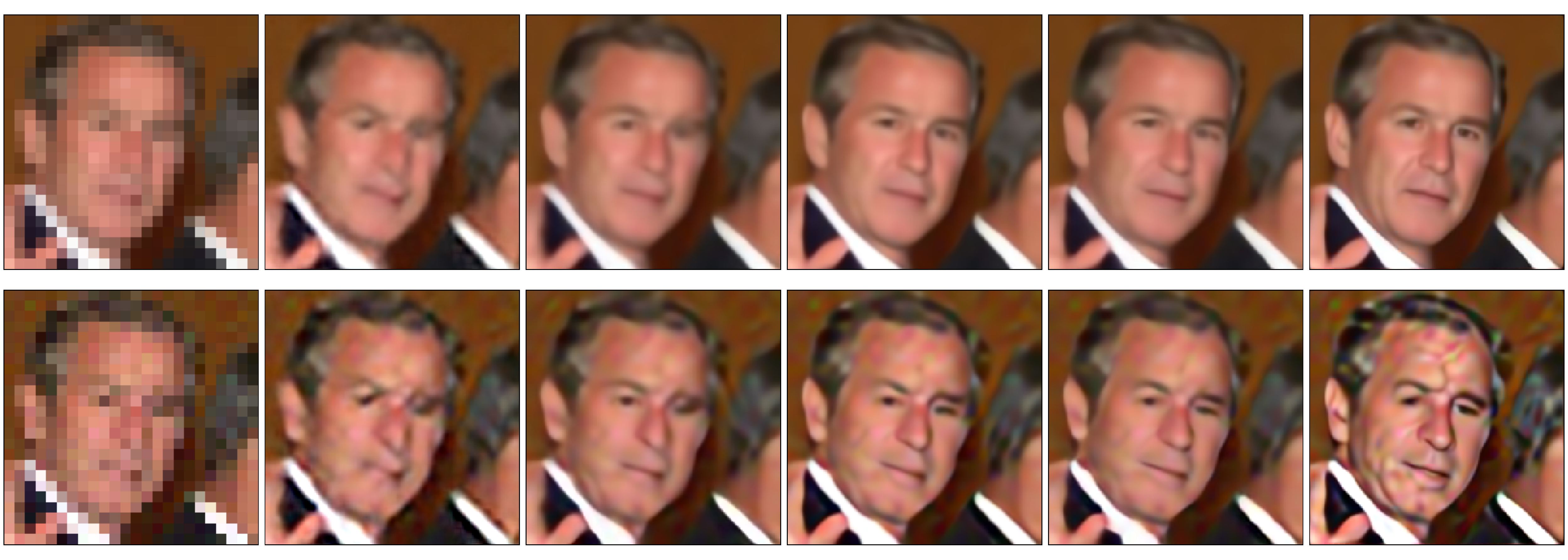}\vspace{1mm}
\end{minipage}
\begin{minipage}{0.03\columnwidth}
\begin{turn}{270}
  \scriptsize{ MS \hspace{8mm} NMS}
  \end{turn}
\end{minipage}
\begin{minipage}{\columnwidth}
\vspace{-1mm}
\scriptsize{\hspace{0.4mm} LR ($24$ px) \hspace{2mm} URDGN \hspace{3mm} LapSRN \hspace{3.2mm} SRResNet \hspace{3.5mm} CARN \hspace{4.8mm} C-SRIP}
\end{minipage}
\caption{Hallucination examples  ($8\times$) 
for the five FH models used in this work (see Sec.~\ref{Sec: Method} for details). 
The top row shows results for a LR image generated with a degradation procedure matching (MS) the one used during training and the bottom row shows results for an image produced by a non-matching degradation function (NMS). Note the difference in the reconstruction quality. 
In this paper, we study the bias introduced into FH models by the training data, which has so far received limited attention in the literature. 
}
\label{fig: sr_bias}\vspace{-3mm}
\end{figure}
Formally, face hallucination is defined as an inverse problem described by the following observation model~\cite{nasrollahi2014super}: 
\begin{equation}
\mathbf{x}=\mathbf{H}\mathbf{y}+\mathbf{n},
\label{Eq: Inverse_SR_problem}
\end{equation}
where $\mathbf{x}$ denotes the observed low-resolution face image, $\mathbf{H}$ stands for a composite down-sampling and blurring operator, $\mathbf{n}$ represents an additive i.i.d. Gaussian noise term with standard deviation $\sigma_n$, and $\mathbf{y}$ is the latent high-resolution face image that needs to be recovered~\cite{nasrollahi2014super}. 
Recent techniques increasingly approach the FH problem in~\eqref{Eq: Inverse_SR_problem} using machine learning methods~\cite{ahn2018carn,bulat2018learn,lai2017lapsrn,nguyen_SR_2018,yu2018face,  yu2016ultra} and try to learn a direct (non-linear) mapping $f_{\theta}$ from the LR inputs to the desired HR outputs, i.e.,
$f_{\theta}: \mathbf{x} \rightarrow \mathbf{y}.$

This mapping is commonly implemented with a parameterized regression model, e.g., a convolutional neural network (CNN), and the parameters of the model $\theta$ are learned through an optimization procedure that minimizes a selected training objective (e.g., an $L_p$ loss) over a set of corresponding LR-HR image pairs. Because the learning procedure is supervised, the image pairs needed for training are constructed by artificially degrading HR training images using a selected degradation function, i.e., a known operator $\mathbf{H}$ and noise level $\sigma_n$. Such an approach ensures that all generated LR images have corresponding HR ground truth faces available for training, but also implicitly defines the type of image degradations the learned model is able to handle. If the actual degradation function encountered with (real-world) test data differs from the one used during training, the result of the face hallucination model may be far from optimal - as  illustrated in Fig.~\ref{fig: sr_bias} for five recent state-of-the-art 
FH models~\cite{ahn2018carn,grm2018face,lai2017lapsrn,ledig2016photo,yu2016ultra}. 

As can be seen from the presented examples, the HR images recovered from a LR input that matches the characteristics of the training data (Fig.~\ref{fig: sr_bias}, top row) are of significantly better quality than those produced from a non-matching LR input image (Fig.~\ref{fig: sr_bias}, bottom row). While all models are able to convincingly ($8\times$) upscale the example $24\times 24$  face  with a matching LR image, the hallucination results exhibit considerable artifacts when a small change in terms of blur and noise is introduced into the degradation procedure. 
These examples show that the bias introduced into the FH models by the training data has a detrimental effect on the quality of the super-resolved faces and may adversely effect the generalization ability of the trained models to data with unseen characteristics. 

Surprisingly, the problem of (face hallucination) model bias has received  little attention from the research community so far. Nevertheless, it has important implications for the generalization abilities of FH models as well as for the performance of high-level vision tasks that rely on the generated hallucination results, most notably face recognition. The existing literature on the generalization abilities of FH techniques is typically focused on generalization across different facial characteristics, such as pose, facial expressions, occlusions or alignment, and less so on the mismatch in the degradation functions used to produce the LR test data or qualitative experiments with real-world imagery. Difficulties with model bias are, therefore, rarely observed. Similarly, when used to improve performance of LR face recognition problems, FH models are predominantly applied on artificially degraded images, leaving the question of generalization to real-world LR data unanswered. 

In this paper, we aim to address these issues and study the problem model bias in the field of face hallucination. We try to answer obvious research questions, such as: How do different image characteristics affect the reconstruction quality of FH models? How do FH models trained on artificially degraded images generalize to real-world data? Do FH models ensure improvements in LR face recognition when applied as a preprocessing step? Are there differences in recognition performance when using either artificially generated or real-world LR data? To answer these and related questions we conduct a rigorous analysis using five recent state-of-the-art FH models and examine in detail: \textit{i)} the mismatch between the degradation procedure used to generate the LR-HR training pairs and the actual degradation function encountered with LR data, \textit{ii)} changes in classifier-independent separability measures before and after the application of FH models, and \textit{iii)} face recognition performance with hallucinated images and a state-of-the-art CNN recognition model. We make interesting findings that point to open and rarely addressed problems in the area of face hallucination 
and provide insights into future research challenges in this area.

\section{Related Work}

\textbf{Bias in computer vision.} Machine learning techniques are known to be sensitive to the characteristics of the training data and typically result in models with sub-optimal generalization abilities if the training data 
is biased towards certain data characteristics.
The 
effect of dataset bias can, for example, be seen in \cite{buolamwini2018gender}, where commercial gender classification systems are shown to have a drop in gender-classification accuracy on darker-skinned subjects compared to lighter-skinned subjects, indicating insufficient training data coverage of the latter. Torralba and Efros~\cite{torralba2011unbiased} dem\-onstrate that image datasets used to train classification models  are heavily biased towards specific appearances of object categories, causing poor performance in cross-dataset experiments. 
Zhao et al. \cite{zhao2017men} show that datasets for semantic role labeling tasks, contain significant gender bias and introduce strong associations between gender labels and  verbs/objects (e.g., \textit{woman} and \textit{cooking}) that lead to biased models for certain labeling tasks. 
These examples show that understanding dataset bias is paramount for the generalization abilities of machine learning models. 
Our work is related to these studies, as we also explore dataset bias. However, different from prior work, we focus on the task of face hallucination, which 
has not been studied from this perspective so far.

\textbf{Face hallucination for face recognition.} Face recognition performance with LR images tends to degrade severely in comparison to HR face data.
To mitigate this problem, a significant body of work resorts to FH models and tries to up-sample images during pre-processing~\cite{farrugia2017face,gunturk2003eigenface, lin2007super,su2016supervised}
or to devise models that jointly learn an upscaling function and recognition procedure~\cite{hennings2008simultaneous,jian2015simultaneous, wu2016deep}. 
While performance improvements are reported with these works, experiments are commonly limited to artificially down-sampled images, findings are then simply extrapolated to real-world data and potential issues due to dataset bias are often overlooked.  

Experiments with real LR images, on the other hand, are scarce in the literature and the usefulness of FH models for face recognition with real-world LR imagery has not received much attention by the research community. As part of our analysis, we study this issue and explore the effect on FH models on data separability and recognition performance on artificially down-sampled and real-world LR data.

\section{Methodology}\label{Sec: Method}

\subsection{Experimental setup}

We conduct our analysis with several  state-of-the-art FH models and LR images of size $24\times 24$ pixels. Since there is no clear distinction on what constitutes a LR image, we select the LR image data to be smaller than $32\times 32$ pixels, which represents an image size, below which most computer vision models are known to deteriorate quickly in performance~\cite{grm2017strengths, torralba200880,wang2016studying}. Given this rather small size, we use an upscaling factor of $8\times$  with the FH models and generate $192\times 192$ images that are  used as the basis for our analysis.

\subsection{Face hallucination (FH) models}

Using the presented setup, we study the effect of dataset bias using five recent FH (or super-resolution) models, i.e.: the Ultra Resolving Discriminative Generative Network (URDGN,~\cite{yu2016ultra}), the Deep Laplacian Super-Resolution Network (LapSRN,~\cite{lai2017lapsrn}),   the Super-Resolution Residual Network (SRResNet,~\cite{ledig2016photo}), the Cascading Residual Network (CARN,~\cite{ahn2018carn}), and the Cascading Super Resolution Network with  Identity Priors (C-SRIP,~\cite{grm2018face}). The selected models differ in the network architecture and training objective, but are all considered to produce state-of-the-art hallucination results as shown in Fig.~\ref{fig: sr_bias}. We also include an interpolation-based method in the experiments to have a baseline for comparisons. A short summary of the models is given below:
\begin{itemize}[leftmargin=*]\vspace{-0.5mm}
    \item \textbf{Bicubic interpolation}~\cite{bicubic} is a learning-free approach that up-samples images by interpolating missing pixel values using Lagrange polynomials, cubic splines, or other similar functions. Unlike  FH models, it does not rely on domain knowledge when generating HR faces. \vspace{-1mm}
    \item \textbf{URDGN} consists of a generator and a discriminator network, and is trained using the generative adversarial network (GAN~\cite{goodfellow2014generative}) framework, where the discriminator is trained to tell apart real and generated HR images, whereas the generator is trained to minimize an $L_2$ reconstruction loss and the accuracy of the discriminator.\vspace{-1mm}
   \item \textbf{LapSRN} represents a CNN-based  model that progressively up-samples LR images by factors of $2$ through bilinear deconvolution and relies on a feature prediction branch to calculate the high-frequency residuals at each scale. Because of the progressive up-sampling, multi-scale supervision signals are used during training. \vspace{-1mm}
   \item \textbf{SRResNet} is a variant of the SRGAN~\cite{ledig2016photo} model that incorporates many of the recent tweaks used in CNN-based super-resolution, such as adversarial training, pixel shuffle up-sampling, batch normalization and leaky ReLU activations. SRResNet represents the generator network of SRGAN trained with the $L_2$ loss. \vspace{-1mm}
  \item \textbf{CARN} consists of a light-weight CNN, which is able to achieve state-of-the-art performance for the general super-resolution problems using an efficient cascading architecture that combines the design principles of densely connected networks~\cite{huang2017densely} and res nets~\cite{He_2016_CVPR}. We use the variant with local and global cascading connections, as opposed to the lighter variants of the network.\vspace{-1mm}
  \item \textbf{C-SRIP} is a CNN-based FH model that incorporates explicit face identity constraints into the training procedure in addition to the main reconstruction objective. The model has a cascaded architecture that allows it to use supervision signals at multiple scales during training.
\end{itemize}

To incorporate face-specific domain knowledge into the models and ensure a fair comparison, we train all models on the CASIA Webface~\cite{yi2014learning} dataset using $494,414$ images of $10,575$ subjects. We crop the $192\times 192$ central part of the images and generate the HR-LR data pairs for training by  blurring the HR images with a Gaussian kernel of $\sigma_b=\frac{8}{3}$ and then downscaling them $8\times$ using bicubic interpolation.

\begin{figure*}[!tb]
\begin{minipage}{0.02\textwidth}
 \end{minipage}
 \hfill
\begin{minipage}{0.29\textwidth}
  \centering
  \includegraphics[width=1\textwidth]{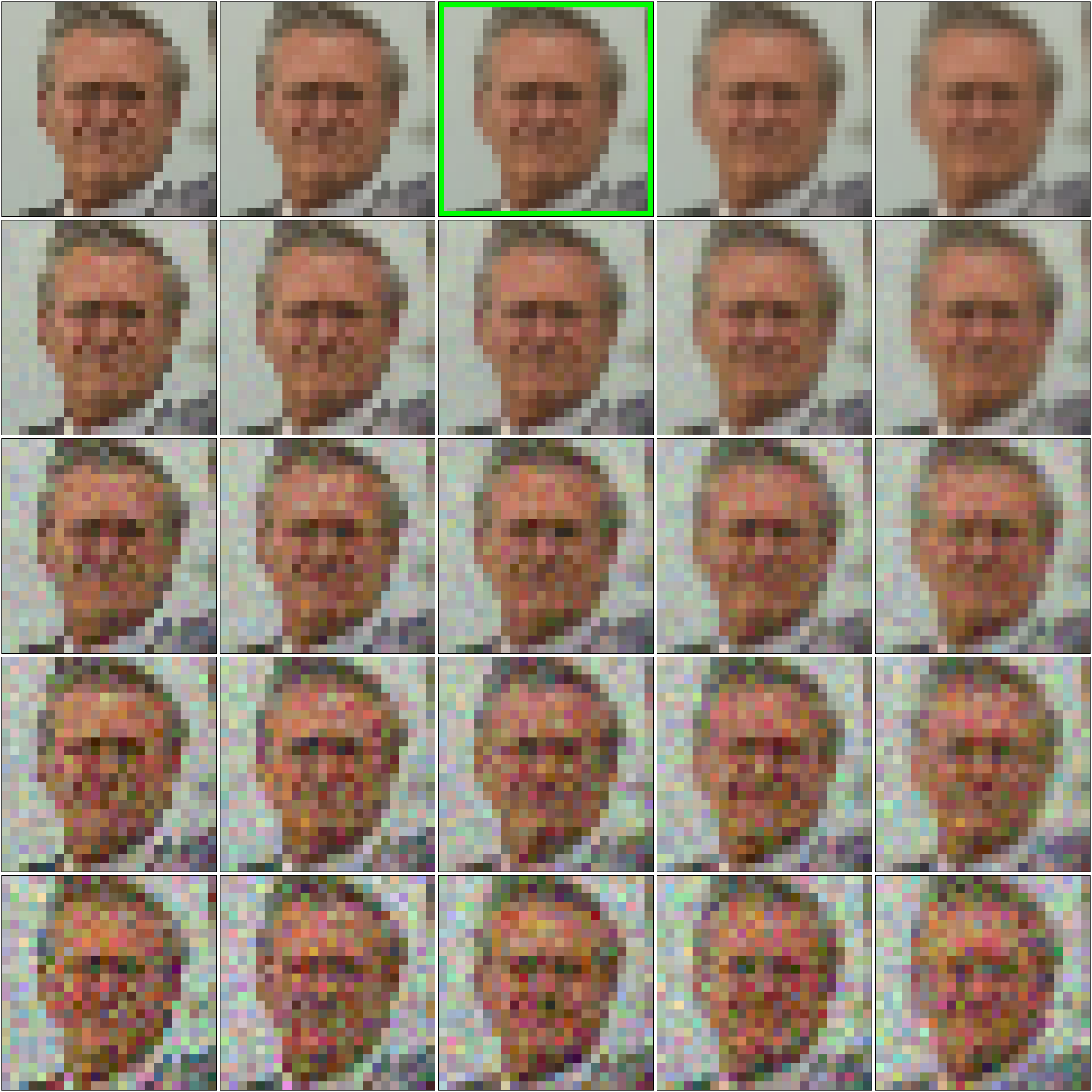}\vspace{0.1mm}
  \text{\footnotesize (a) LR inputs ($\sigma_n$ vs. $\sigma_b$)}
 \end{minipage}
 \hfill
 \begin{minipage}{0.29\textwidth}
  \centering
  \includegraphics[width=1\textwidth]{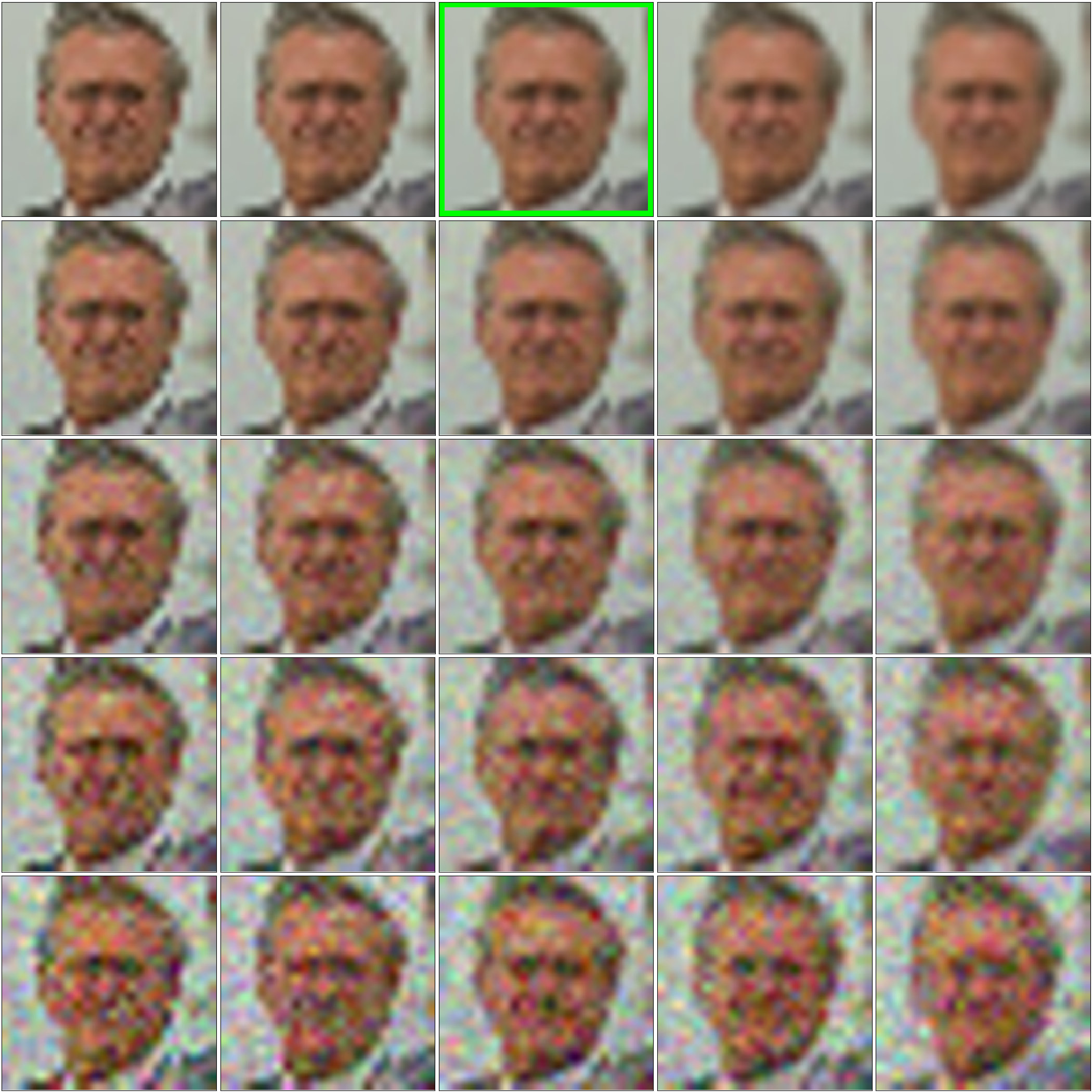}\vspace{0.1mm}
  \text{\footnotesize (b) Bicubic ($\sigma_n$ vs. $\sigma_b$)}
 \end{minipage}
 \hfill
  \begin{minipage}{0.29\textwidth}
  \centering
  \includegraphics[width=1\textwidth]{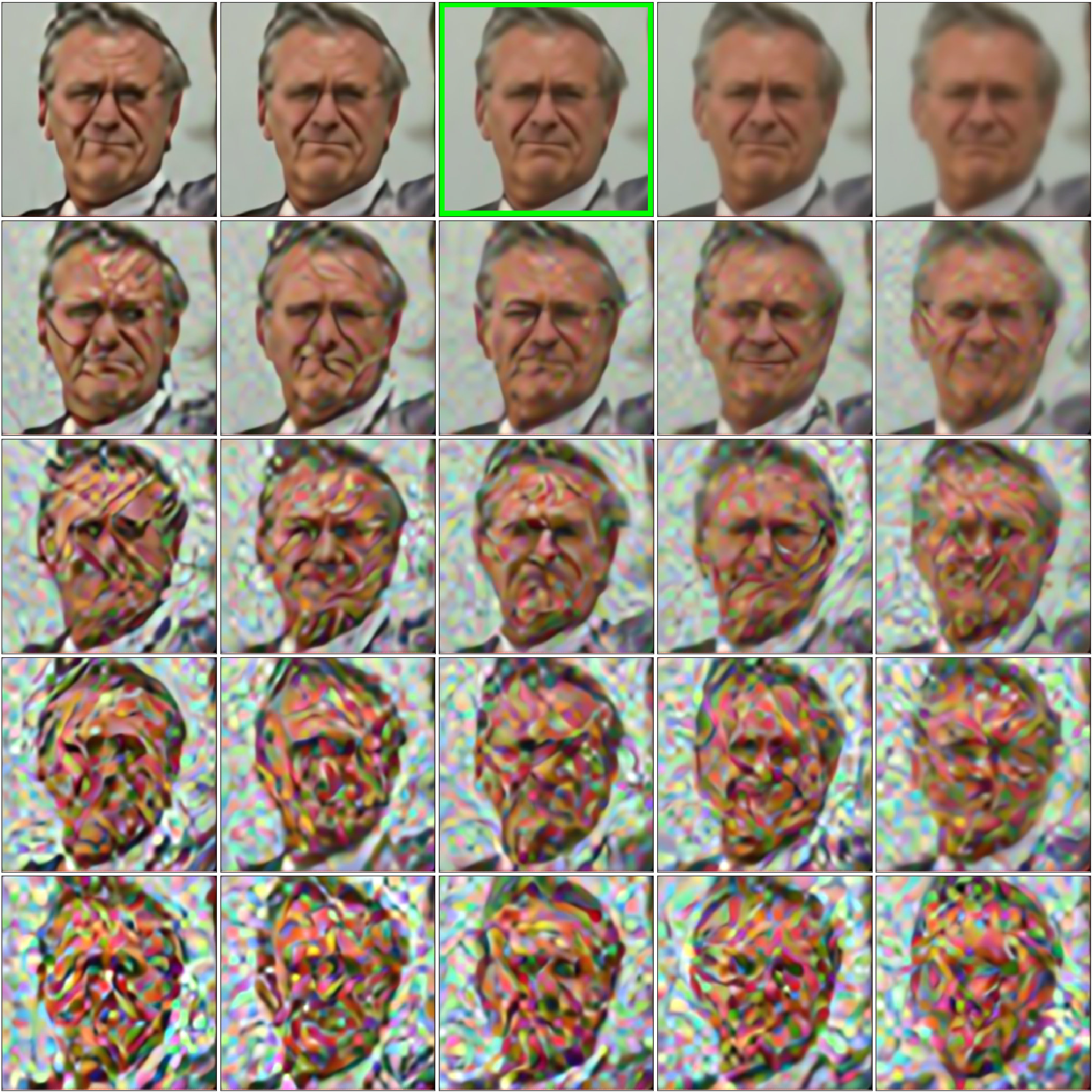}\vspace{0.1mm}
  \text{\footnotesize (f) C-SRIP ($\sigma_n$ vs. $\sigma_b$)}
 \end{minipage}
 \hfill
 \begin{minipage}{0.02\textwidth}
 \end{minipage}\vspace{2mm}
\caption{Reconstruction capabilities of the learning-free bicubic interpolation a selected FH model. The image block on the left (with samples of size $24 \times 24$ pixels) illustrates the effect of increasing noise ($\sigma_n$, increases vertically) and blur ($\sigma_b$, increases horizontally) for a sample LR LFW image, the second and third block show $192 \times 192$ reconstructions generated by bicubic interpolation and C-SRIP, respectively. Images marked green are generated with a degradation function matching the one used during training. For the FH model good HR reconstructions are achieved only with images degraded similarly as the training data, whereas interpolation ensures reasonable reconstructions with all input images. Results for the remaining FH models are shown in the Appendix. Best viewed zoomed in.}\vspace{0mm}

\label{fig:SR_grids_noise}
\end{figure*}

\begin{figure*}[!tb]
 \begin{minipage}{0.162\textwidth}
  \centering
  \includegraphics[width=1\textwidth,trim=6mm 75mm 6mm 6mm, clip]{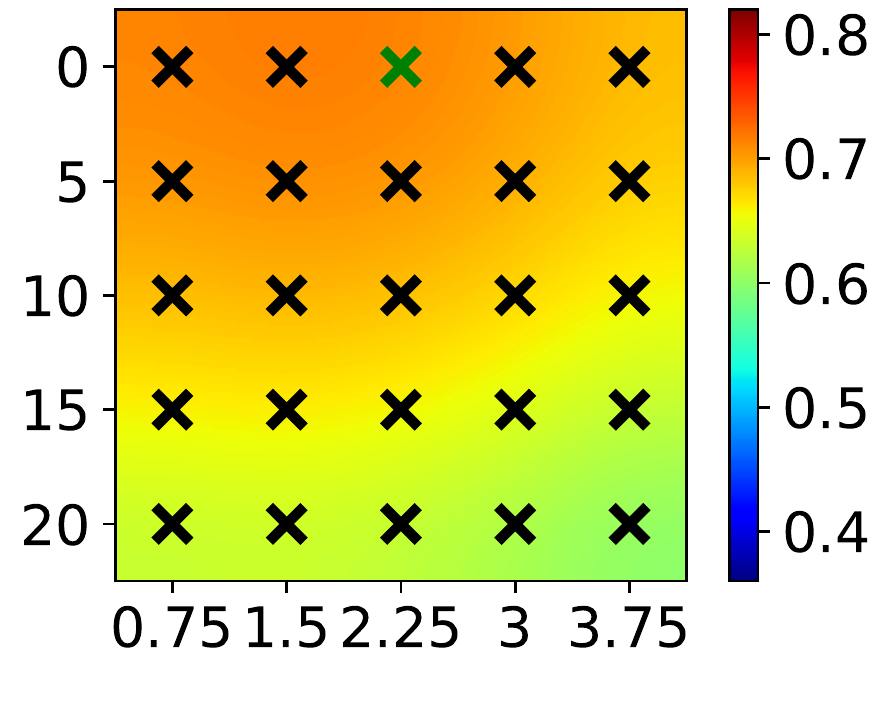}\vspace{-2mm}
  \text{\footnotesize (a) Bicubic} 
 \end{minipage}
 \hfill
  \begin{minipage}{0.162\textwidth}
  \centering
  \includegraphics[width=1\textwidth,trim=6mm 75mm 6mm 6mm, clip]{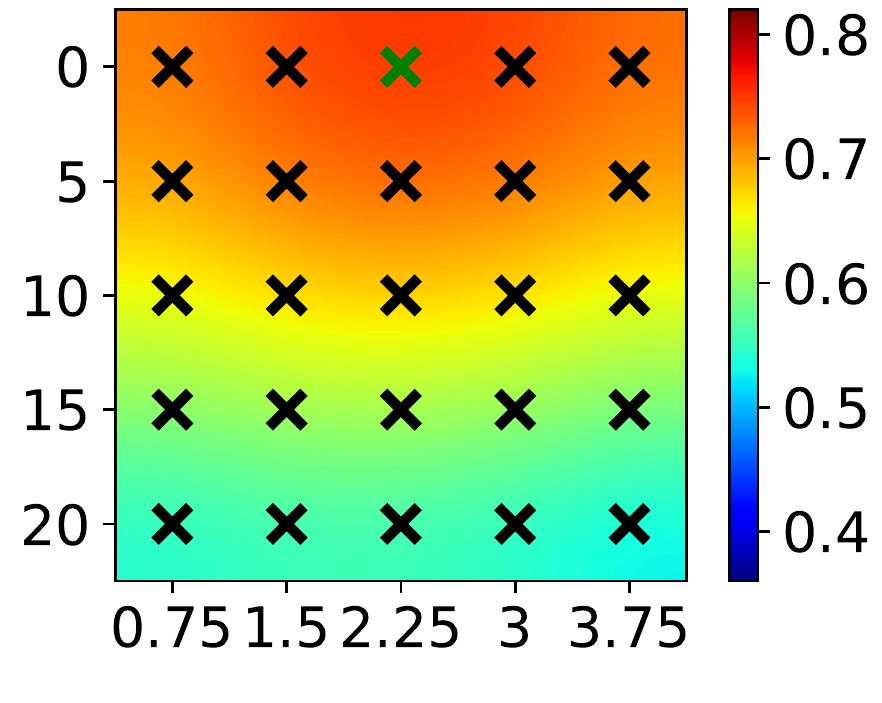}\vspace{-2mm}
  \text{\footnotesize (b) URDGN} 
 \end{minipage}
 \hfill
  \begin{minipage}{0.162\textwidth}
  \centering
  \includegraphics[width=1\textwidth,trim=6mm 75mm 6mm 6mm, clip]{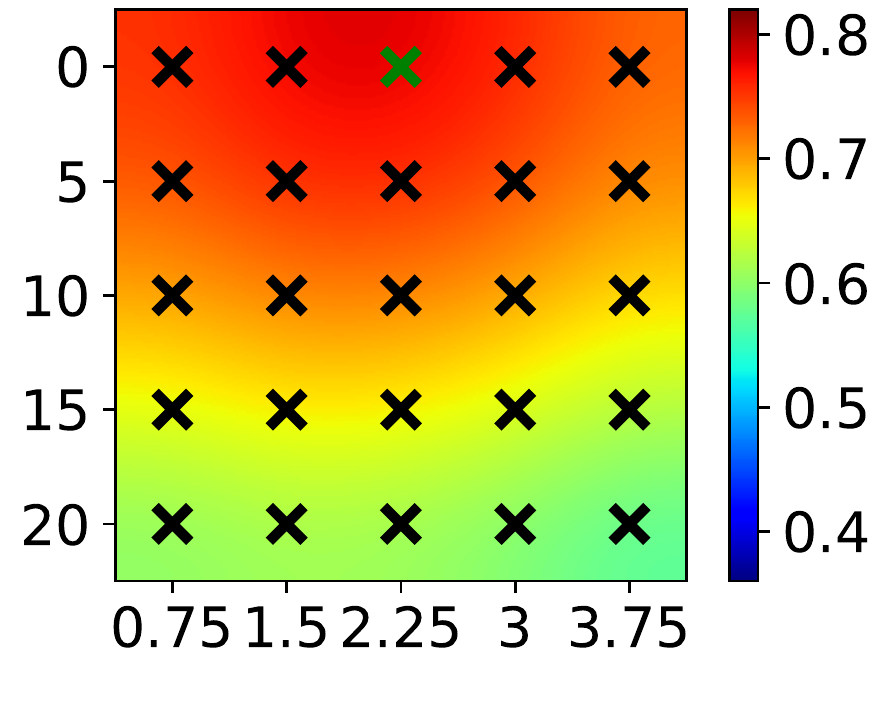}\vspace{-2mm}
  \text{\footnotesize (c) LapSRN} 
 \end{minipage}
 \hfill
  \begin{minipage}{0.162\textwidth}
  \centering
  \includegraphics[width=1\textwidth,trim=6mm 75mm 6mm 6mm, clip]{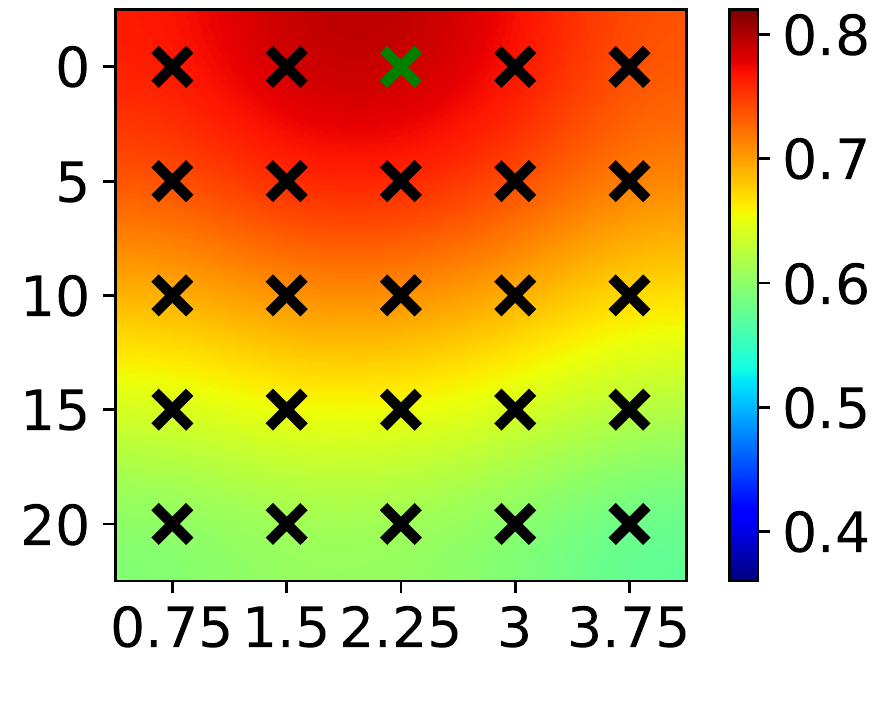}\vspace{-2mm}
  \text{\footnotesize (d) CARN} 
 \end{minipage}
 \hfill
   \begin{minipage}{0.162\textwidth}
  \centering
  \includegraphics[width=1\textwidth,trim=6mm 75mm 6mm 6mm, clip]{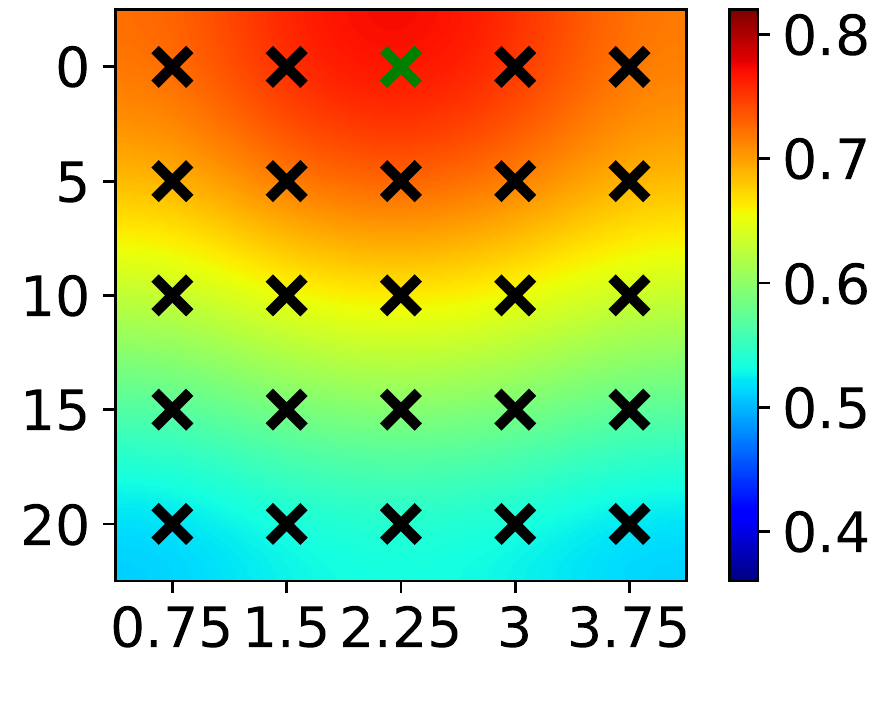}\vspace{-2mm}
  \text{\footnotesize (d) SRResNet} 
 \end{minipage}
 \hfill
   \begin{minipage}{0.162\textwidth}
  \centering
  \includegraphics[width=1\textwidth,trim=6mm 75mm 6mm 6mm, clip]{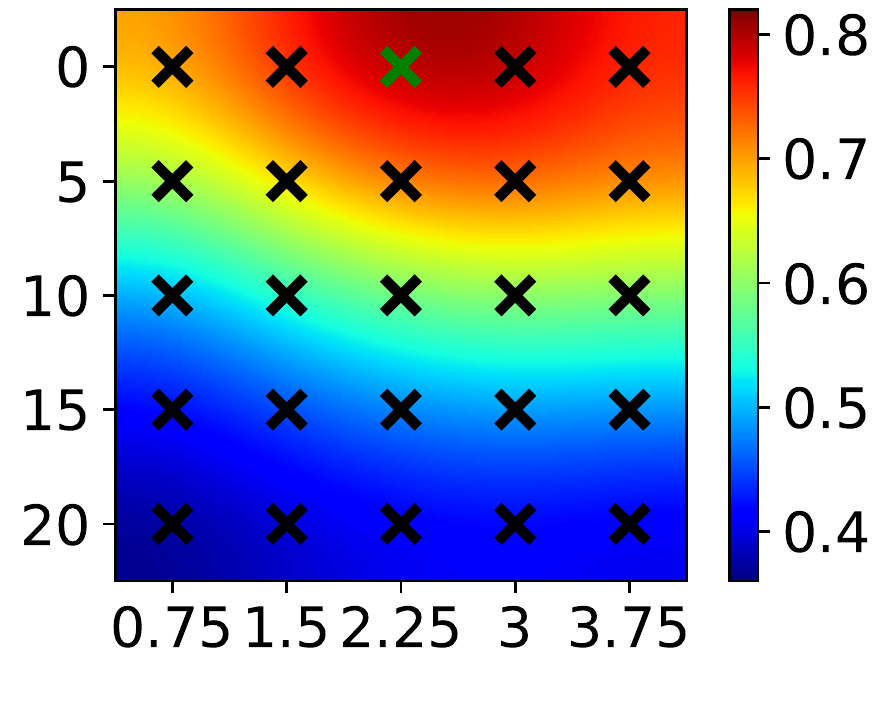}\vspace{-1mm}
  \text{\footnotesize (d) C-SRIP} 
 \end{minipage}\vspace{2mm}
\caption{Reconstruction capabilities with mismatching degradation functions due to different blur and noise levels. The heat maps show the average SSIM values computed over artificially degraded LFW images. The points marked in the heat maps correspond to the sampled levels of noise ($\sigma_n$, increases vertically) and blur  ($\sigma_b$, increases horizontally). The value of $\sigma_n$ and $\sigma_b$ that was used for training is marked green. Note that all FH models achieve good reconstructions only around values that match the training setup. Best viewed in color.}\vspace{-1.5mm}
\label{fig:heat_maps}
\end{figure*}

\subsection{Datasets.} \label{subsec:datasets}

We conduct experiments on the Labeled Face in the Wild (LFW~\cite{huang2007labeled}) and SCFace~\cite{grgic2011scface} datasets. We introduce artificial down-sampling to simulate  low image resolutions with LFW and use the SCFace images to explore the effect of training data bias on real-world LR images.
\begin{itemize}[leftmargin=*]\vspace{-0.5mm}
    \item \textbf{LFW} is one of the most popular face dataset available, mainly due to the unconstrained settings in which the images were captured. The dataset~\cite{huang2007labeled} consists of $13,233$ face images of size $250\times 250$ pixels belonging to $5749$ subjects. For the experiments, we use only the central crop of the images to have faces of similar proportion to the ones used during FH model training. \vspace{-1mm}
    \item \textbf{SCface} contains images of $130$  subjects that are split between a gallery set, containing $130$ high-resolution frontal mug\-shots ($1$ per subject), and a larger probe set of surveillance-camera images. The daylight camera set, which we use for our experiments, consists of images from $5$ different security cameras. 
Each subject is recorded by each camera at $3$ different distances, resulting in a total of $130\times 5\times 3 = 1950$ probe set images. We crop facial areas from all images based on the provided facial landmarks prior to the experiments.
\end{itemize}

\subsection{Bias exploration with synthetic LR data}

We start our analysis by exploring the sensitivity of FH models to a controlled mismatch in the degradation function. We first crop the ($192\times 192$) central part of the LFW images and generate baseline LR test data using the same degradation function as used during training. To simulate the mismatch, we generate additional sets of LR data from LFW by varying the standard deviations of the Gaussian blurring kernel $\sigma_b$ and Gaussian noise term $\sigma_n$, which define $\mathbf{H}$ and $\mathbf{n}$ in~\eqref{Eq: Inverse_SR_problem}. We consider five different values for each parameter and select $\sigma_b$ from $[0.75, 1.5, 2.25, 3, 3.75]$ and $\sigma_n$ from $[0, 5, 10, 15, 20]$. Because the LR test data is generated artificially, the HR ground truth can be used to evaluate the reconstruction capabilities of the FH models for each combination of $\sigma_b$ and $\sigma_n$. Note that it is in general infeasible to include all possible data variations in the training procedure, so there will always be image characteristics that have not been accounted for by data augmentation. The selected noise and blur levels are therefore as reasonable factors as any to simulate the mismatch. 

From the hallucination examples in Fig.~\ref{fig:SR_grids_noise} we see that visually convincing results for the FH model are produced only for LR images generated with blur and noise levels similar to those used during training (close to the images marked green), and deteriorate quickly as the difference to the training blur and noise levels gets larger (see Appendix for additional results). The interpolation baseline produces less convincing results compared to the best hallucinated image of C-SRIP, but  does also introduces lesser distortions with images of other blur and noise levels. A similar observation can also be made for the remaining FH models based on the results in Fig.~\ref{fig:heat_maps}, where average structural similarity (SSIM) values computed over the entire LFW dataset are shown for different levels of noise and blur. Here, the computed SSIM scores are shown in the form of interpolated heat maps for all five FH models and the baseline (bicubic) interpolation procedure. 
The first thing to notice is that the degradation in reconstruction quality is also visible for the (learning-free) interpolation method. This suggests that the reconstruction problem gets harder with increased noise and blur levels 
and the worsened reconstruction quality is not linked exclusively to the mismatch in the degradation function. However, the heat maps also clearly show that performance degrades much faster for the FH models than for the interpolation approach and that the degradation is particularly extreme for the C-SRIP model, which otherwise results in the highest peak SSIM score among all models. 

In general, all FH models achieve significantly higher SSIM scores with matching degradation functions (see  green point in Fig.~\ref{fig:heat_maps}) than the interpolation approach, but their performance falls below bicubic interpolation at the highest noise and blur levels - see lowest heat map part in Fig.~\ref{fig:heat_maps}.
This is an important finding and implies that for imaging conditions that are difficult to model  and challenging to reproduce using \eqref{Eq: Inverse_SR_problem}, interpolation may still be a better choice for recovering HR faces than FH models, which require representative HR-LR image pairs for training.

The presented results are consistent with recent studies~\cite{stutz2018disentangling,su2018robustness}, which suggest that the performance of CNN models may come at the expense of robustness and that trying to learn models that are more robust to varying imaging conditions leads to less accurate results. We observe similar behaviour with the tested FH models (compare the heat maps of C-SRIP and URDGN, for example) and hypothesize that the relatively narrow focus of the models on specific degradation functions may be one of the reasons for the convincing performance of recent CNN-based FH models. 

\subsection{Bias exploration with synthetic and real data}

Next, we explore the impact of dataset bias with synthetic LR images from LFW and with real-world surveillance data from SCFace, where the observed image degradations due to the acquisition hardware are not well modelled by the training degradation function. Since there is no HR ground truth available for the SCFace data, measuring the reconstruction quality is not possible with this dataset. We therefore focus on face recognition, which is regularly advocated in the literature as one of the main applications for FH models~\cite{farrugia2017face,gunturk2003eigenface,lin2007super,su2016supervised}, and use it as a proxy for face hallucination performance. 
Because this task is different from the reconstruction task studied above, we first run experiments with artificially degraded LFW images  to have a baseline for later comparisons with results obtained on real-world SCFace data. We note that recognition experiments add another dimension to our analysis, as we now also explore the impact of the dataset bias on the semantic content of the reconstructed HR data and not only on the perceived quality of the hallucinated faces. 

For the experiments, we use a ResNet-$101$ model~\cite{He_2016_CVPR} and train it for face recognition on a dataset of close to $1.8$ images and $2622$ identities~\cite{VGGface}. We select the mdel because of its state-of-the-art performance~\cite{masi2016we,ranjan2017hyperface} and that fact that an open-source implementation is readily available. We perform network surgery on the trained ResNet-$101$ and use the activations from the penultimate network layer as a $512$-dimensional descriptor of the input face images. 

\begin{figure}[t!]
\centering\vspace{-0.7mm}
\begin{minipage}{0.51\columnwidth}
\centering
  \includegraphics[width=\textwidth]{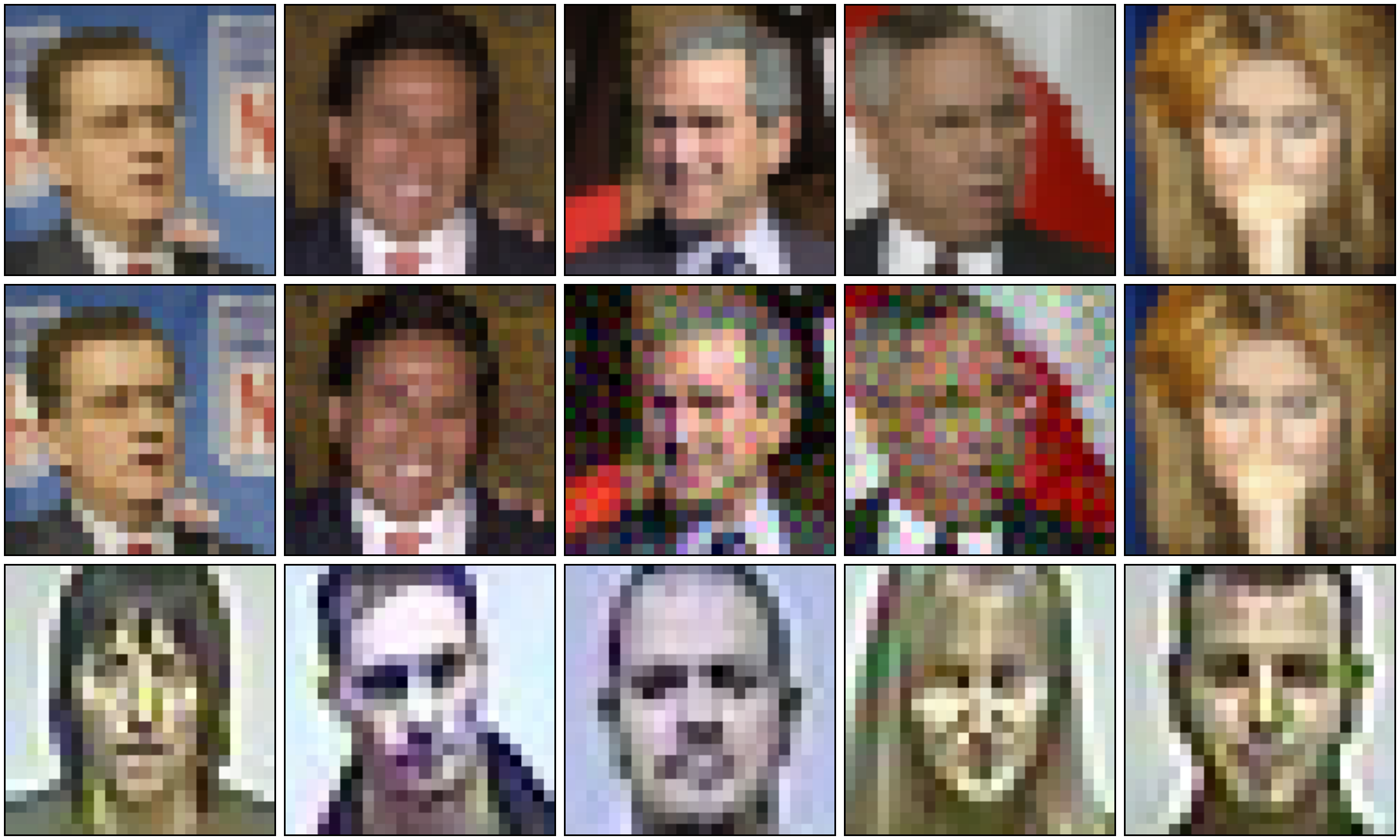}
\end{minipage}
\begin{minipage}{0.48\columnwidth}
\centering
  \includegraphics[width=\textwidth]{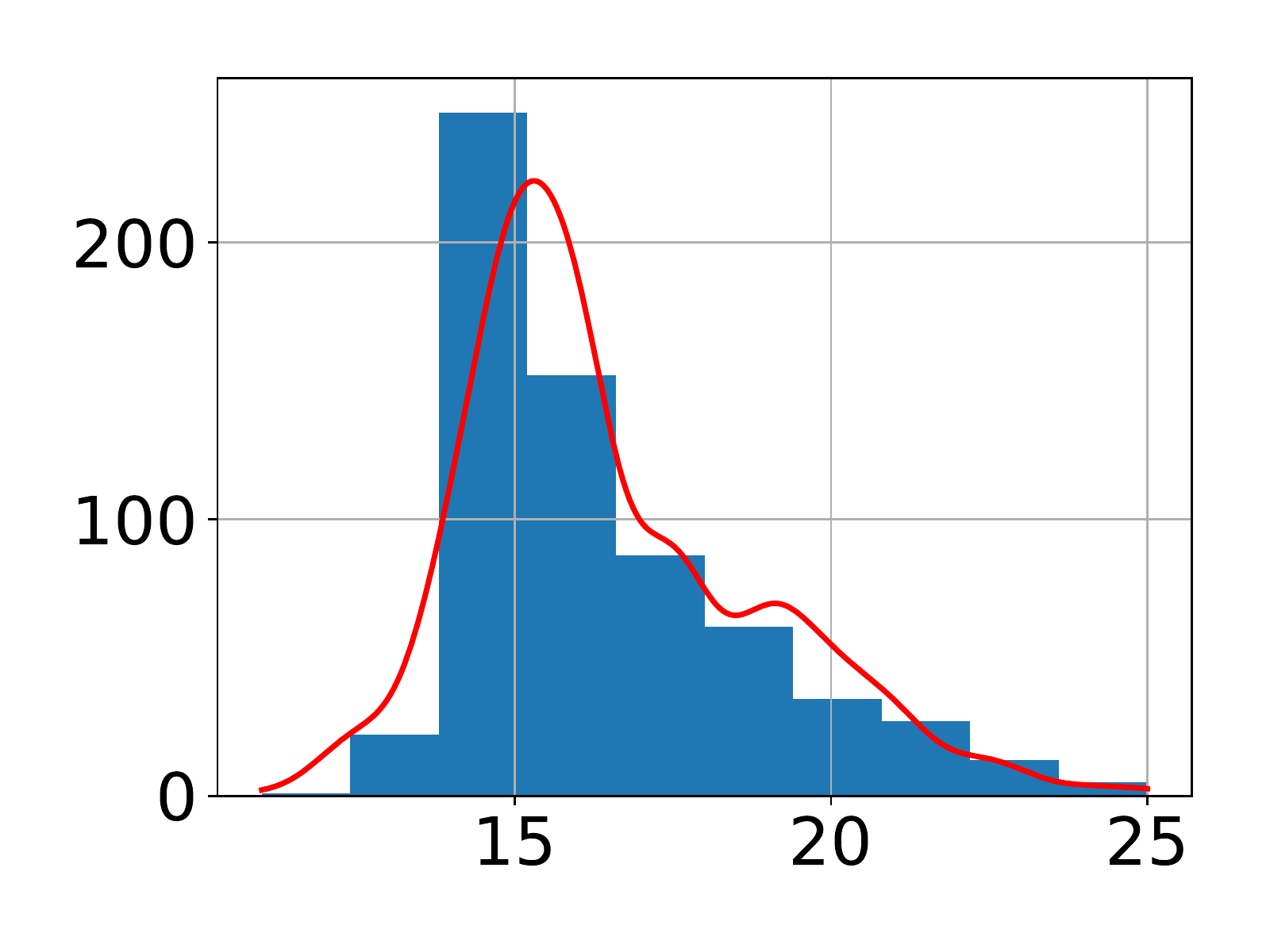}
\end{minipage}
 \caption{Examples of LR LFW and SCFace images used in the experiments. Left: the first row shows LFW samples degraded using the \textit{matching} scheme (MS), the next row shows LFW images degraded with the \textit{non-matching} scheme (NMS) and the last row shows images from SCFace. Right: distribution of SCFace image widths/heights (in $px$) for faces captured at the largest distance.\vspace{-2.5mm}}
 \label{fig:face_samples_scface_lfw}
\end{figure}
\begin{figure*}
\centering\vspace{-0.5mm}
 \begin{minipage}{0.239\textwidth}\centering
  \includegraphics[width=\textwidth]{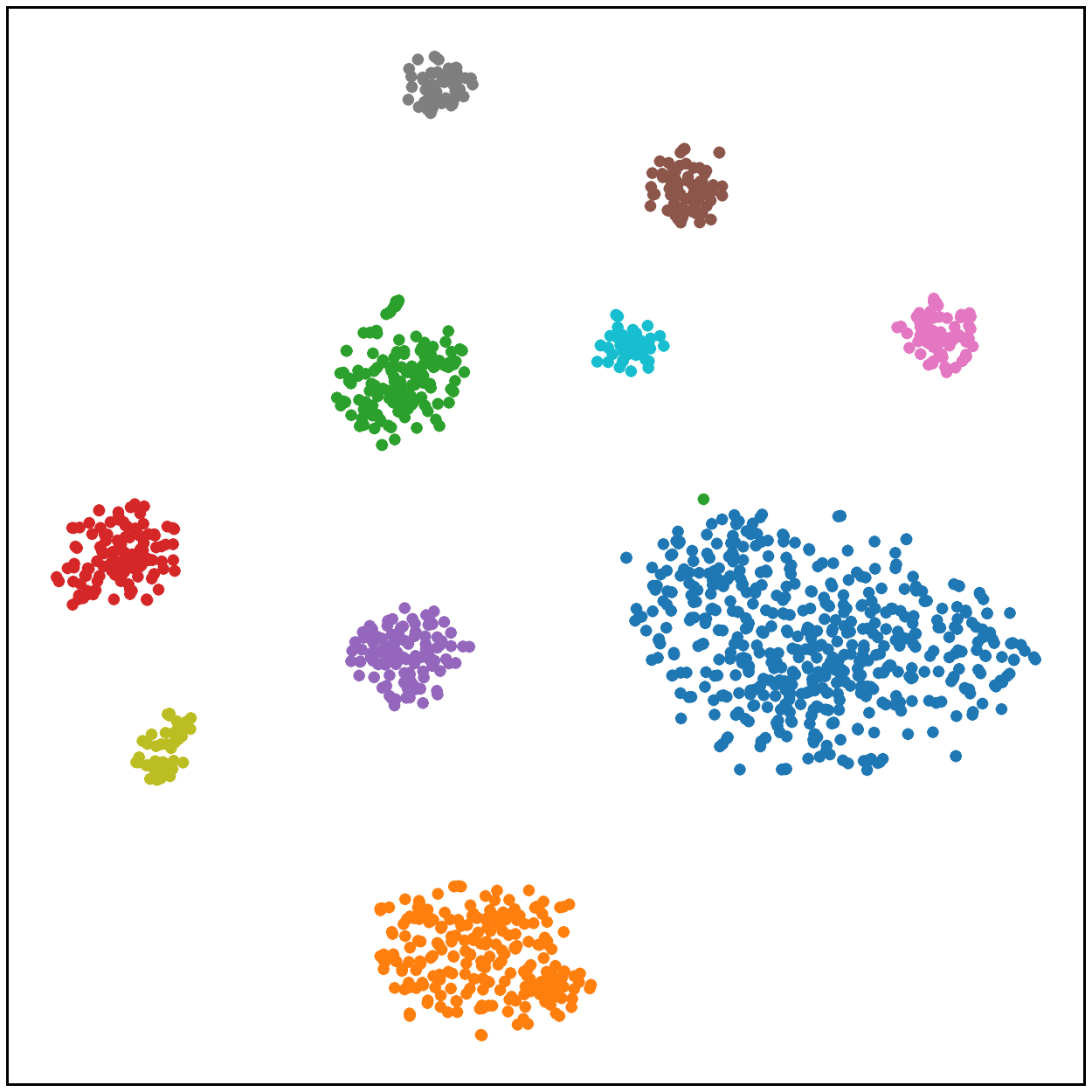}\vspace{0.3mm}
  \text{\footnotesize (a) HR images}
 \end{minipage}
  \begin{minipage}{0.118\textwidth}\centering
  \includegraphics[width=\textwidth]{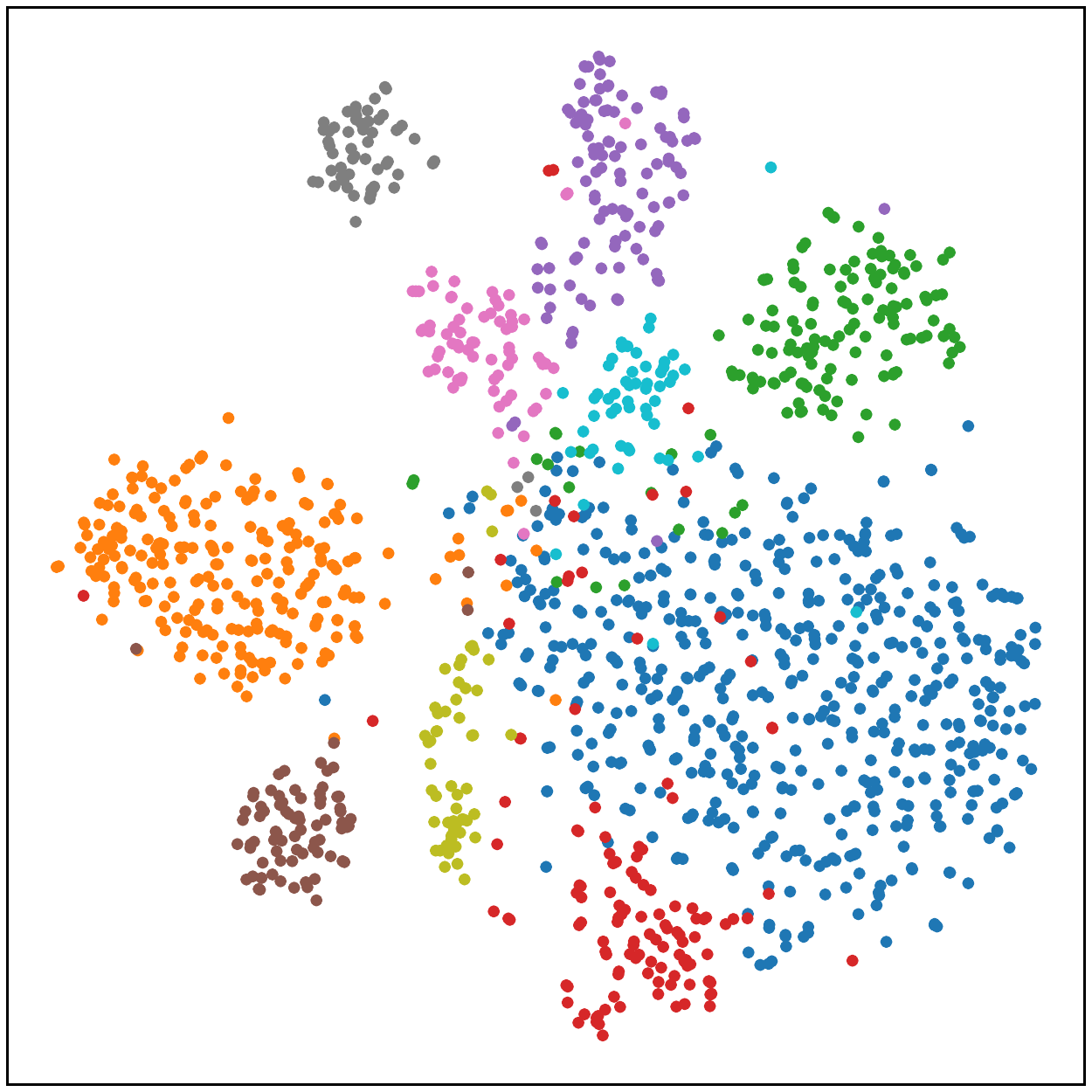}
  \includegraphics[width=\textwidth]{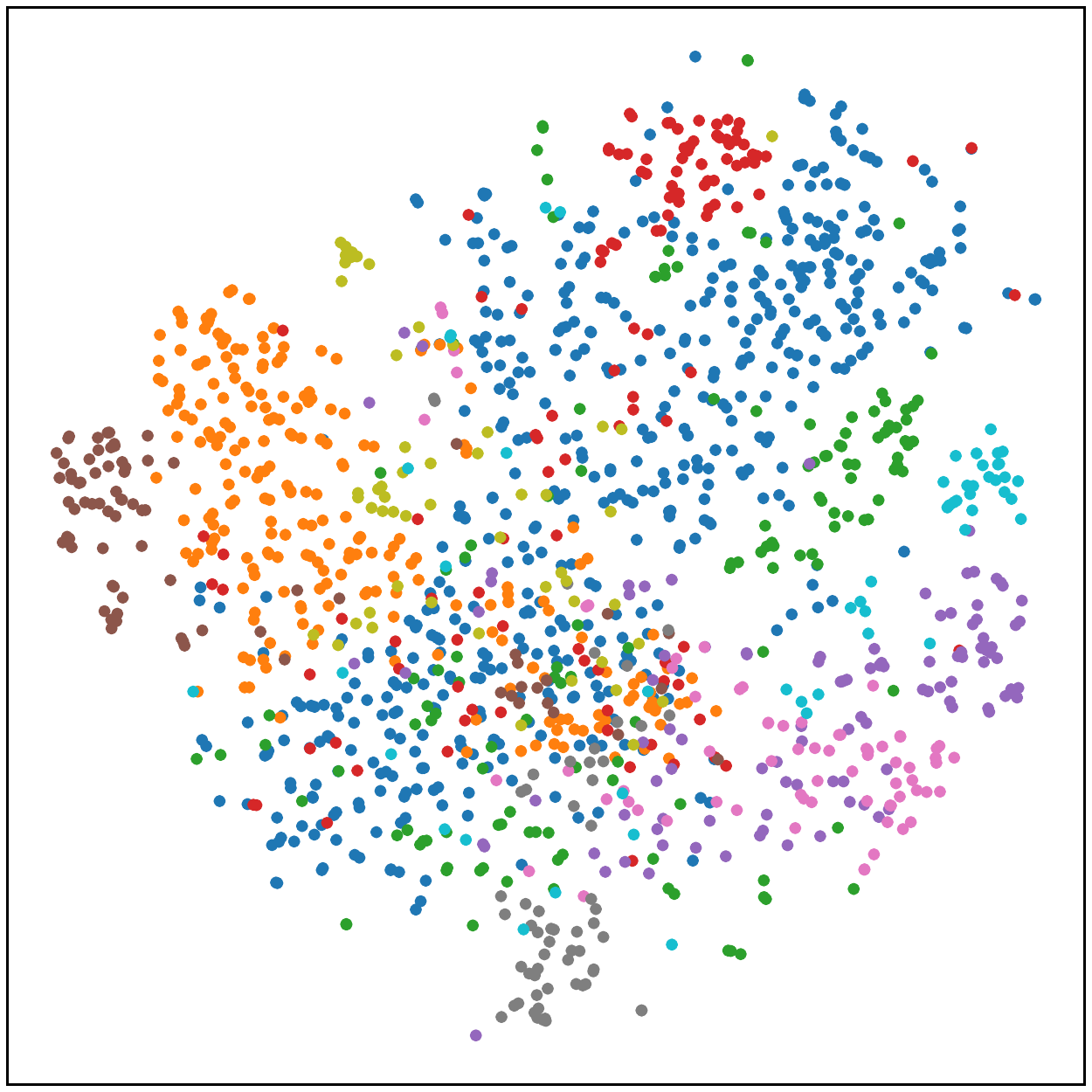}\vspace{0.3mm}
  \text{\footnotesize (b) Bicubic}
 \end{minipage}
   \begin{minipage}{0.118\textwidth}\centering
  \includegraphics[width=\textwidth]{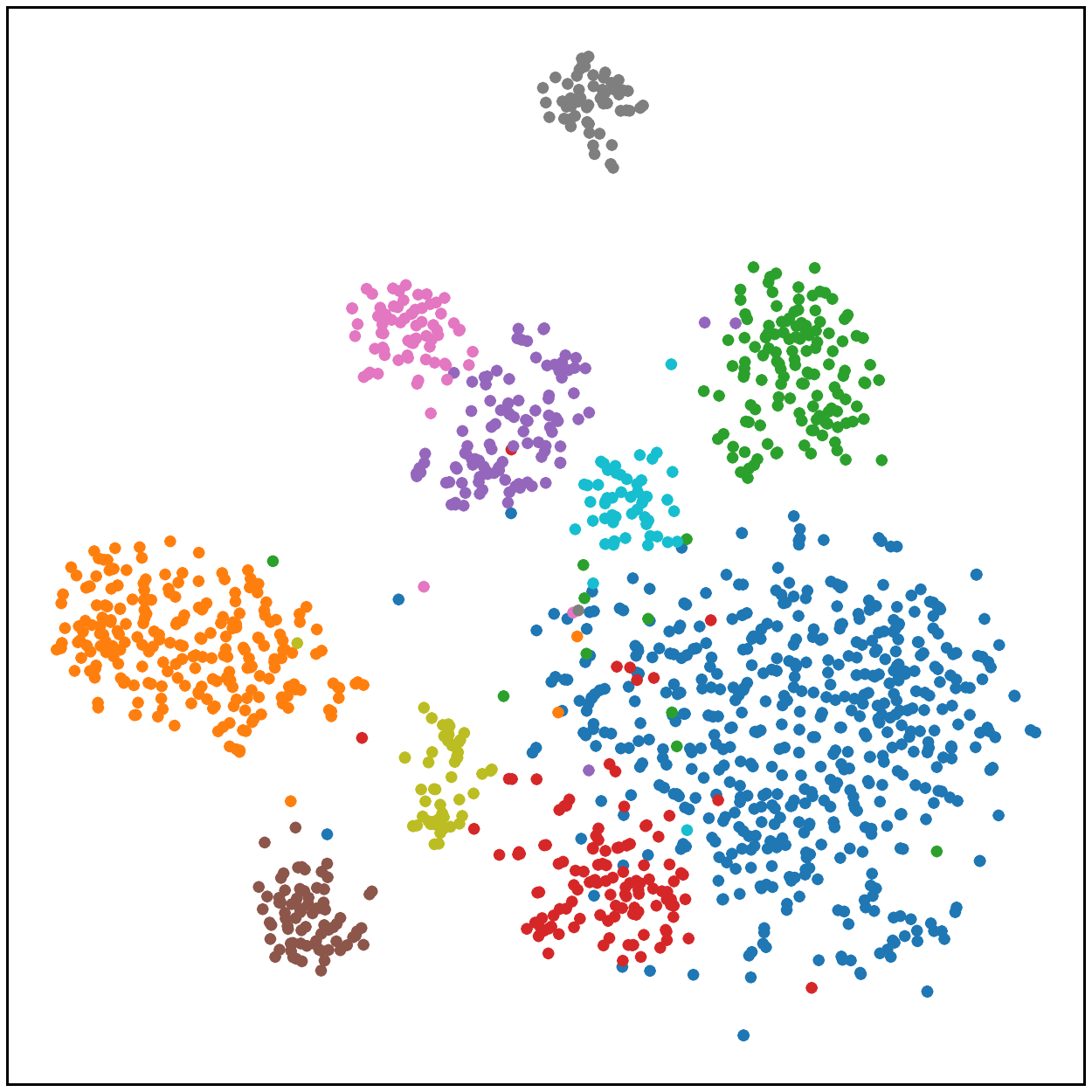}
  \includegraphics[width=\textwidth]{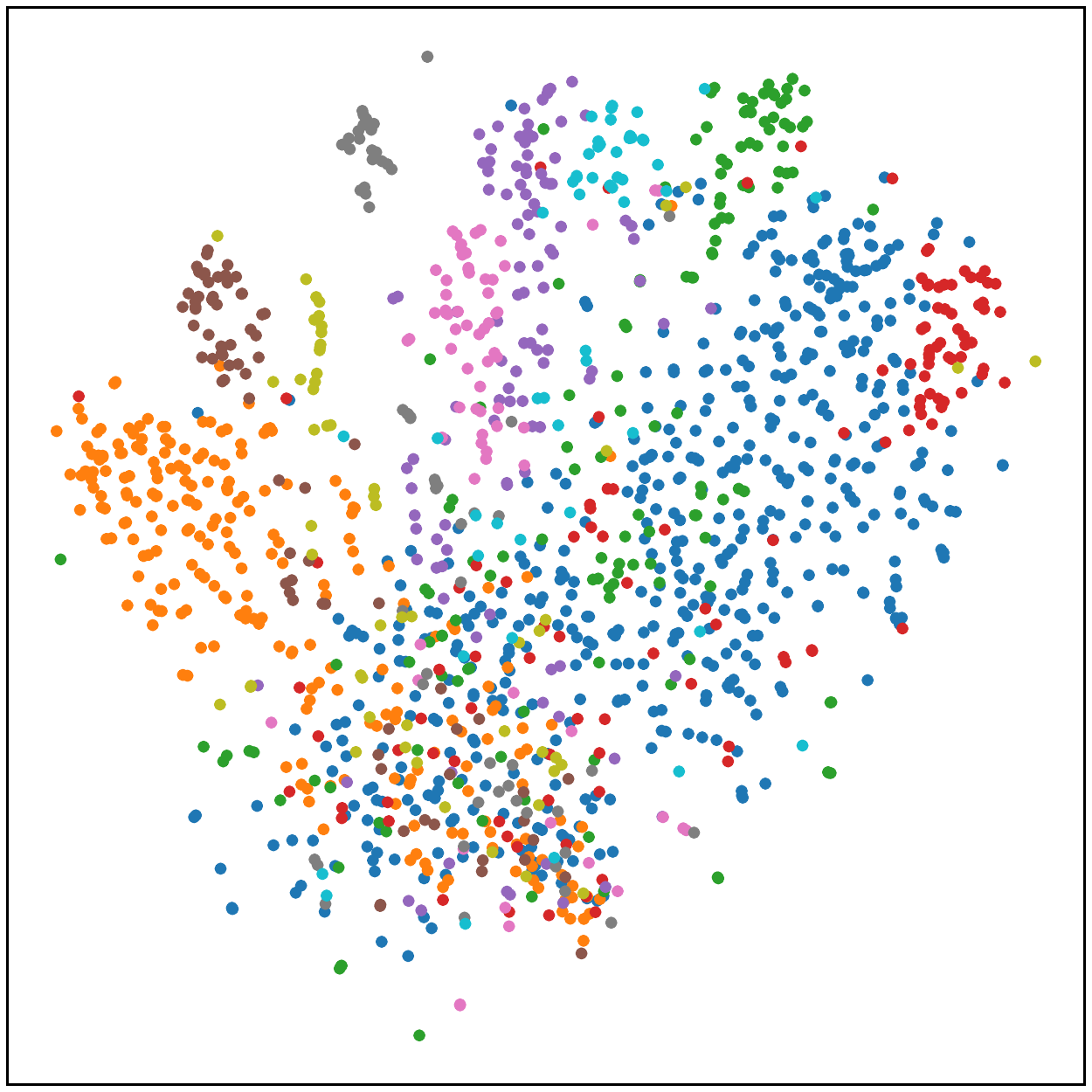}\vspace{0.3mm}
  \text{\footnotesize (c) URDGN}
 \end{minipage}
   \begin{minipage}{0.118\textwidth}\centering
  \includegraphics[width=\textwidth]{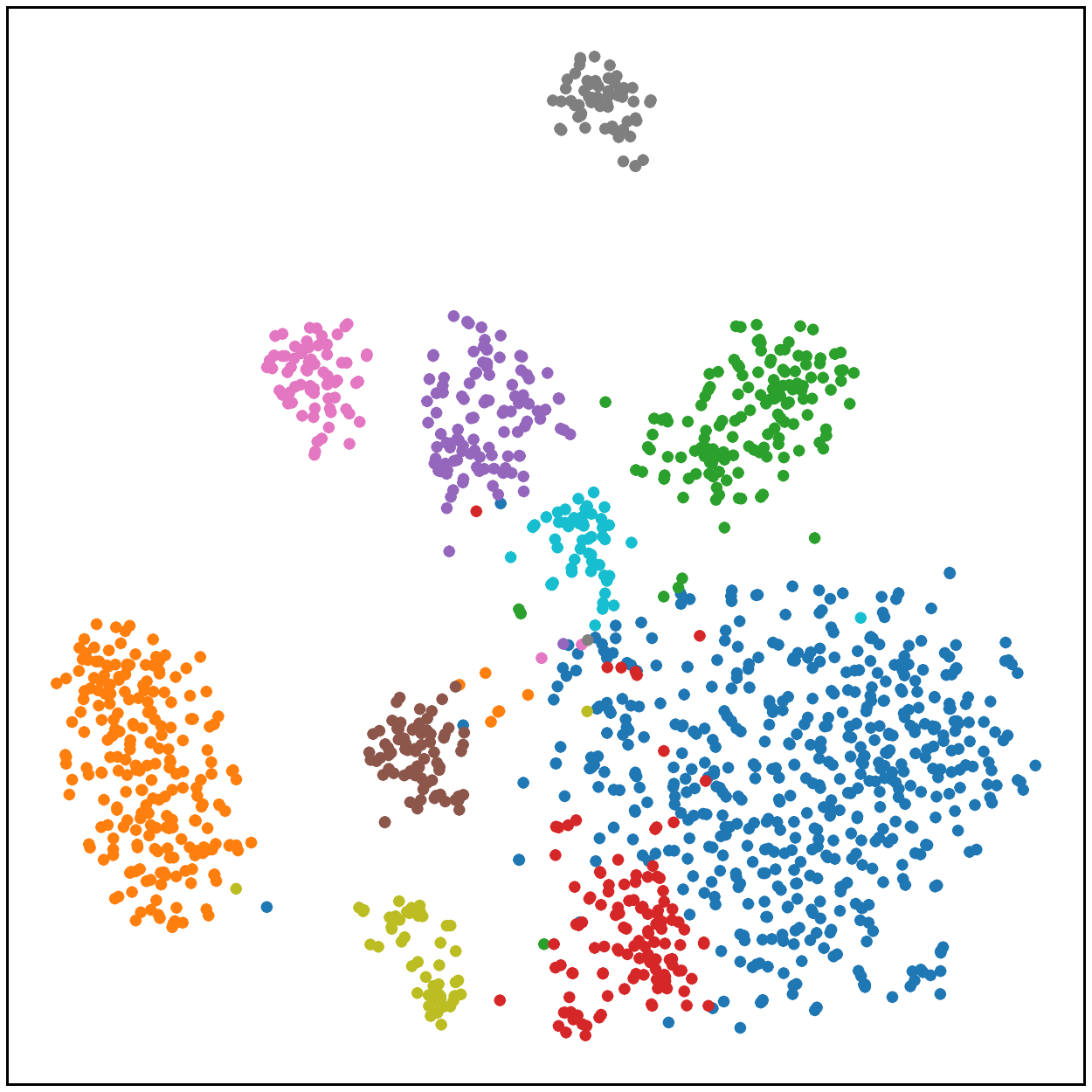}
  \includegraphics[width=\textwidth]{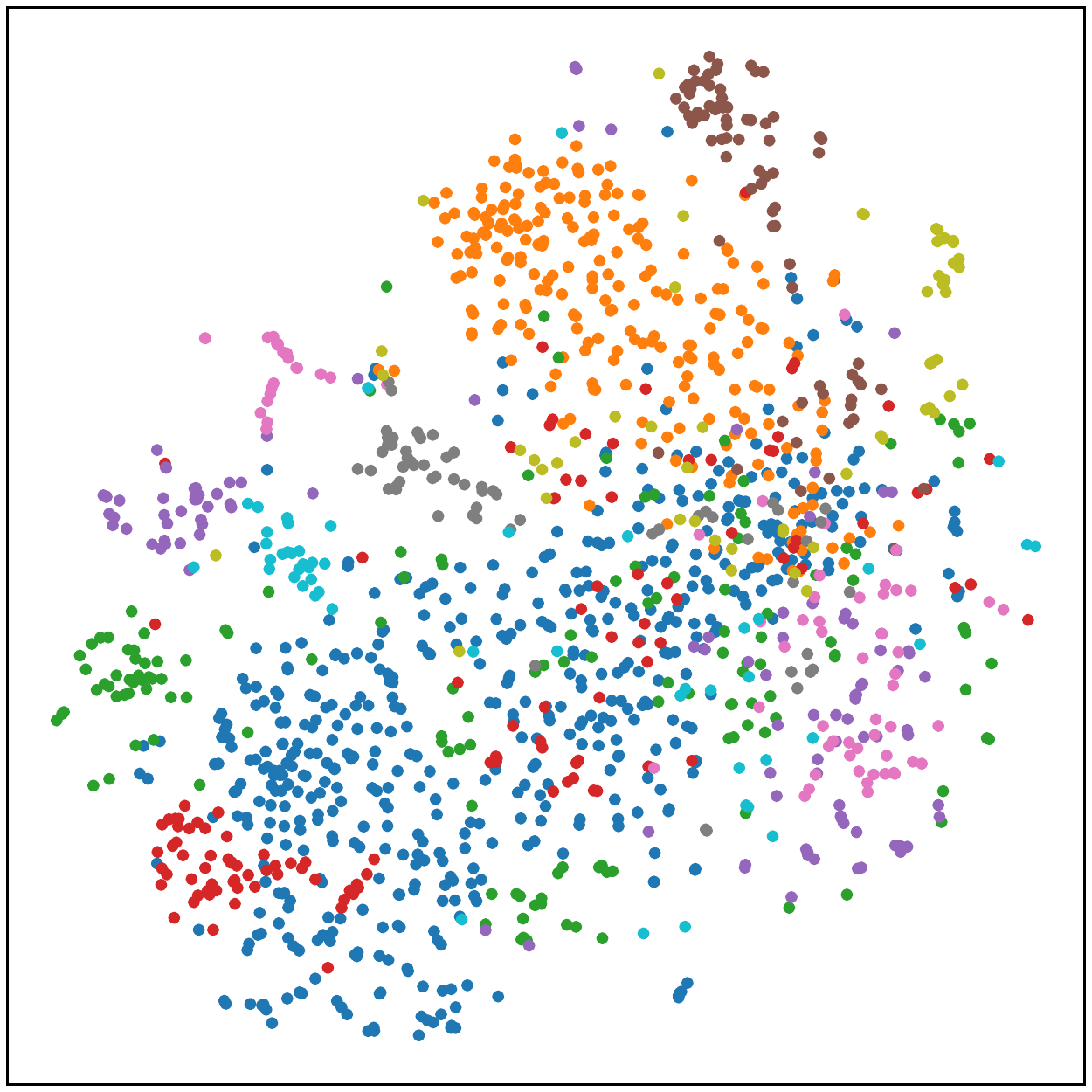}\vspace{0.3mm}
  \text{\footnotesize (d) LapSRN}
 \end{minipage}
   \begin{minipage}{0.118\textwidth}\centering
  \includegraphics[width=\textwidth]{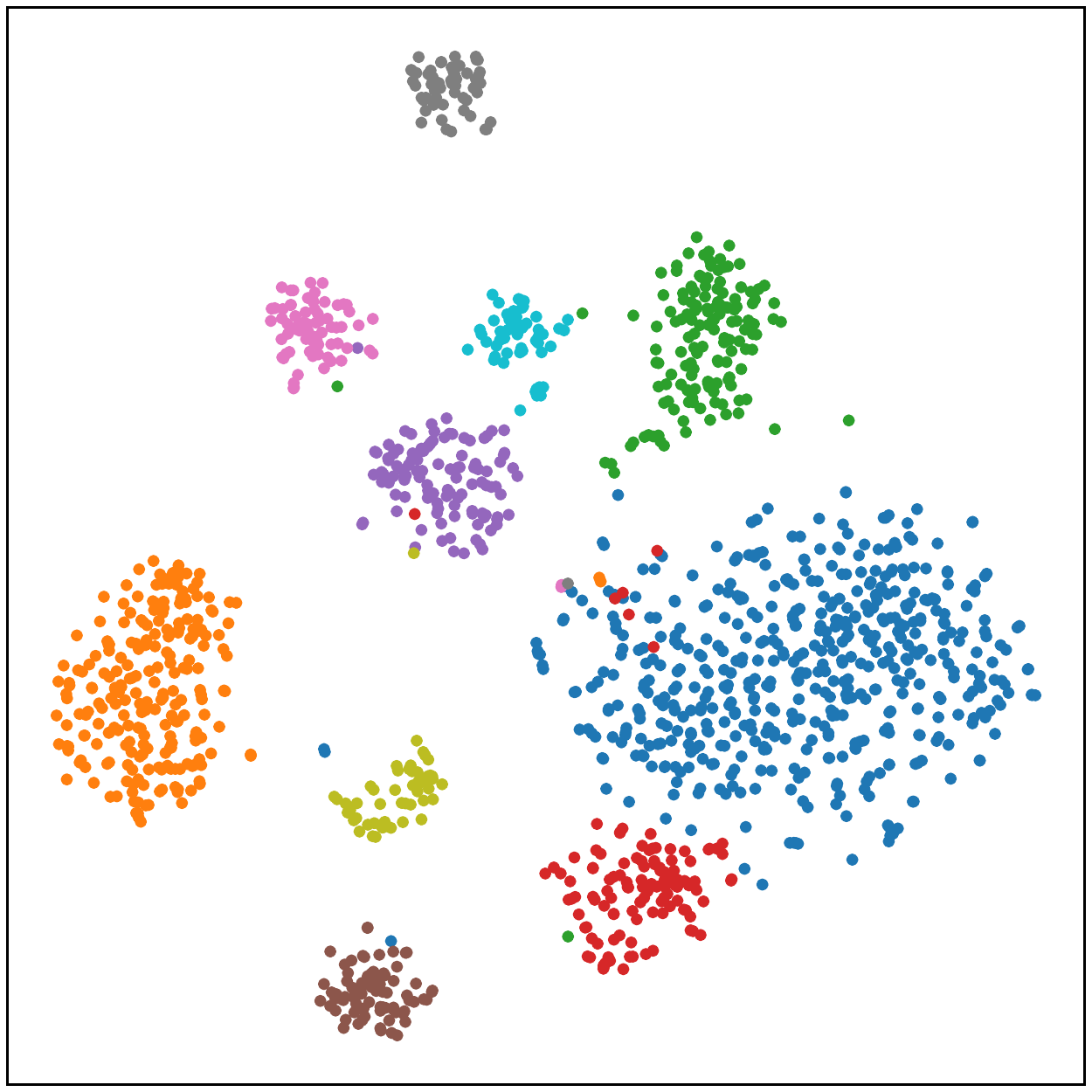}
  \includegraphics[width=\textwidth]{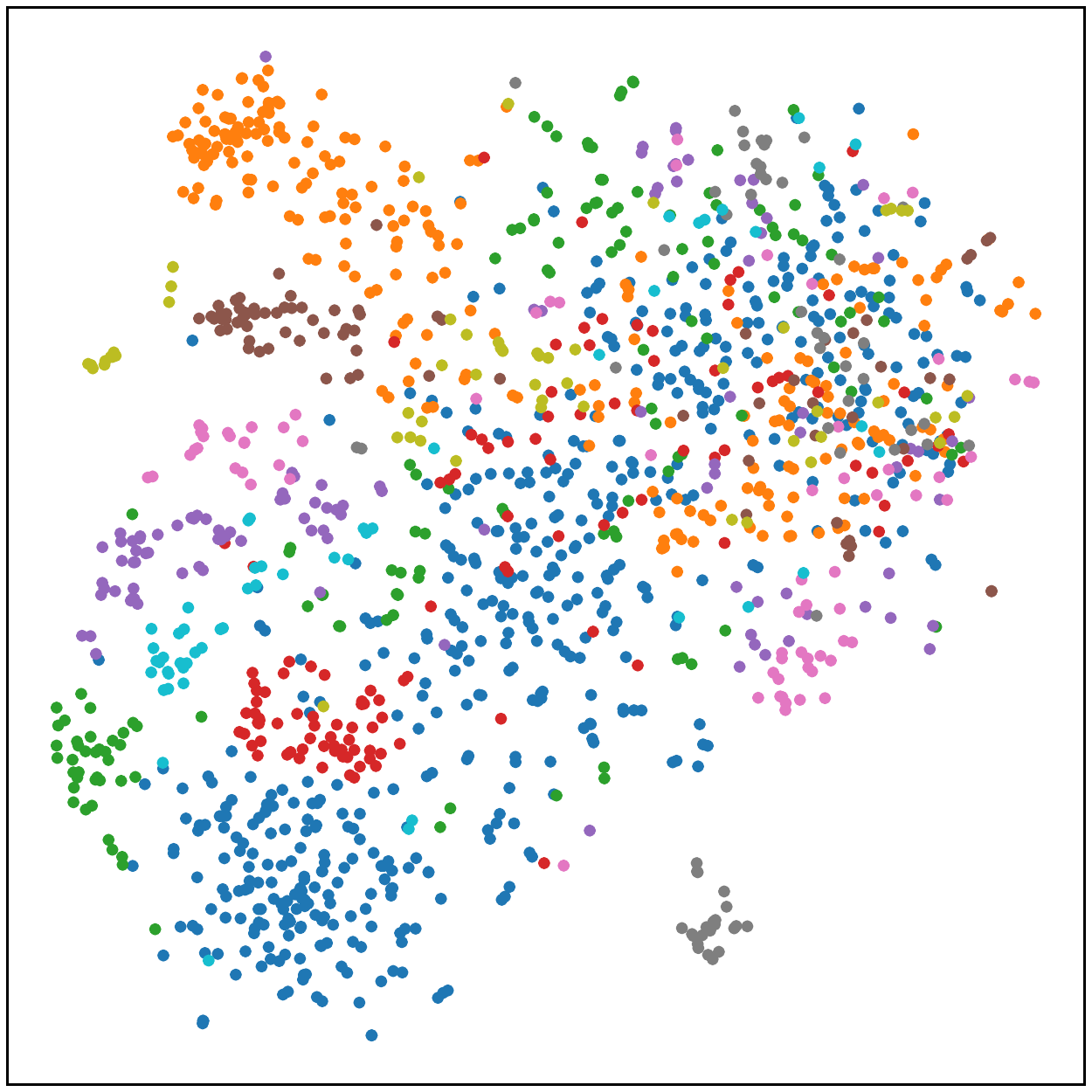}\vspace{0.3mm}
  \text{\footnotesize (e) CARN}
 \end{minipage}
   \begin{minipage}{0.118\textwidth}\centering
  \includegraphics[width=\textwidth]{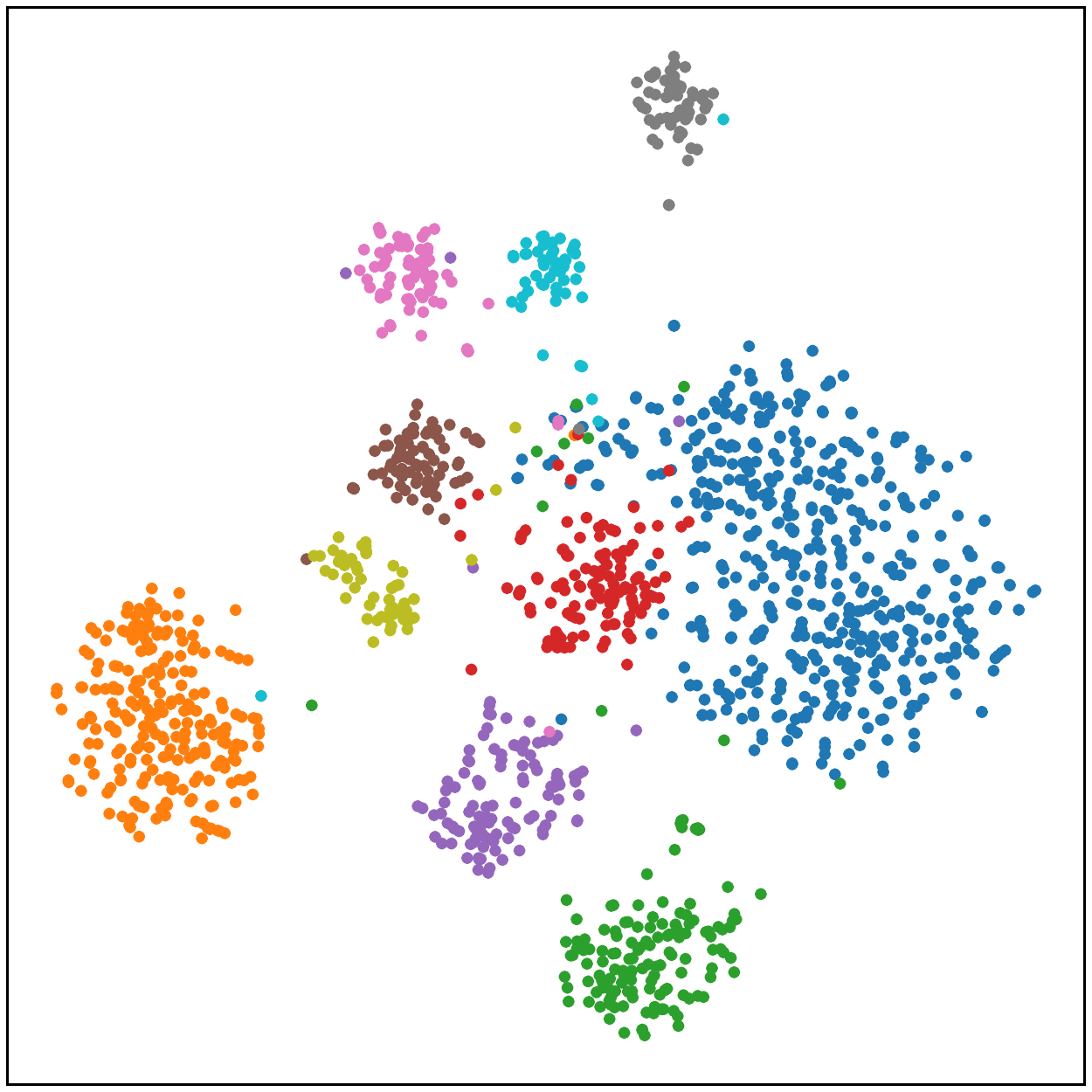}
  \includegraphics[width=\textwidth]{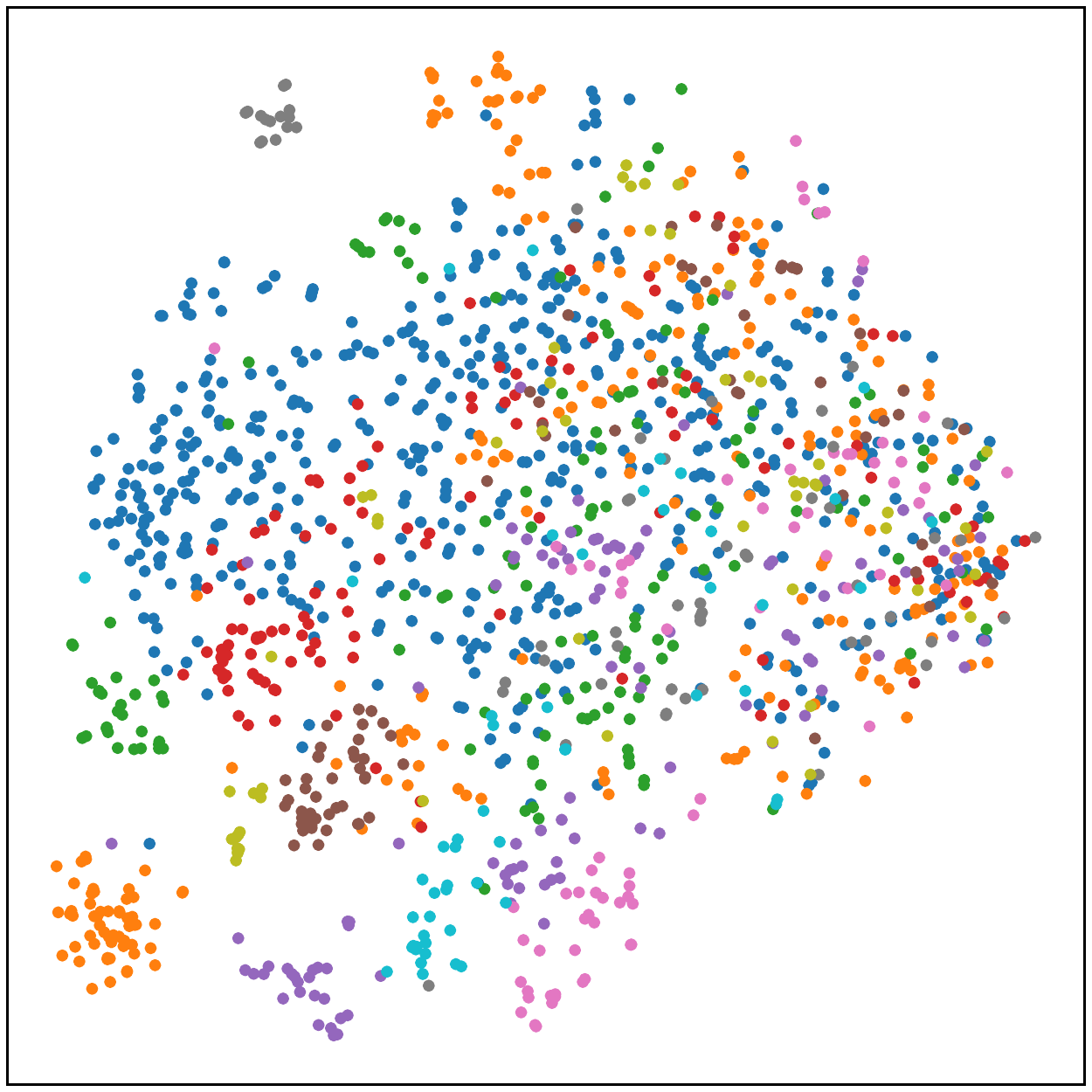}\vspace{0.3mm}
  \text{\footnotesize (f) SRResNet}
 \end{minipage}
   \begin{minipage}{0.118\textwidth}\centering
  \includegraphics[width=1\textwidth]{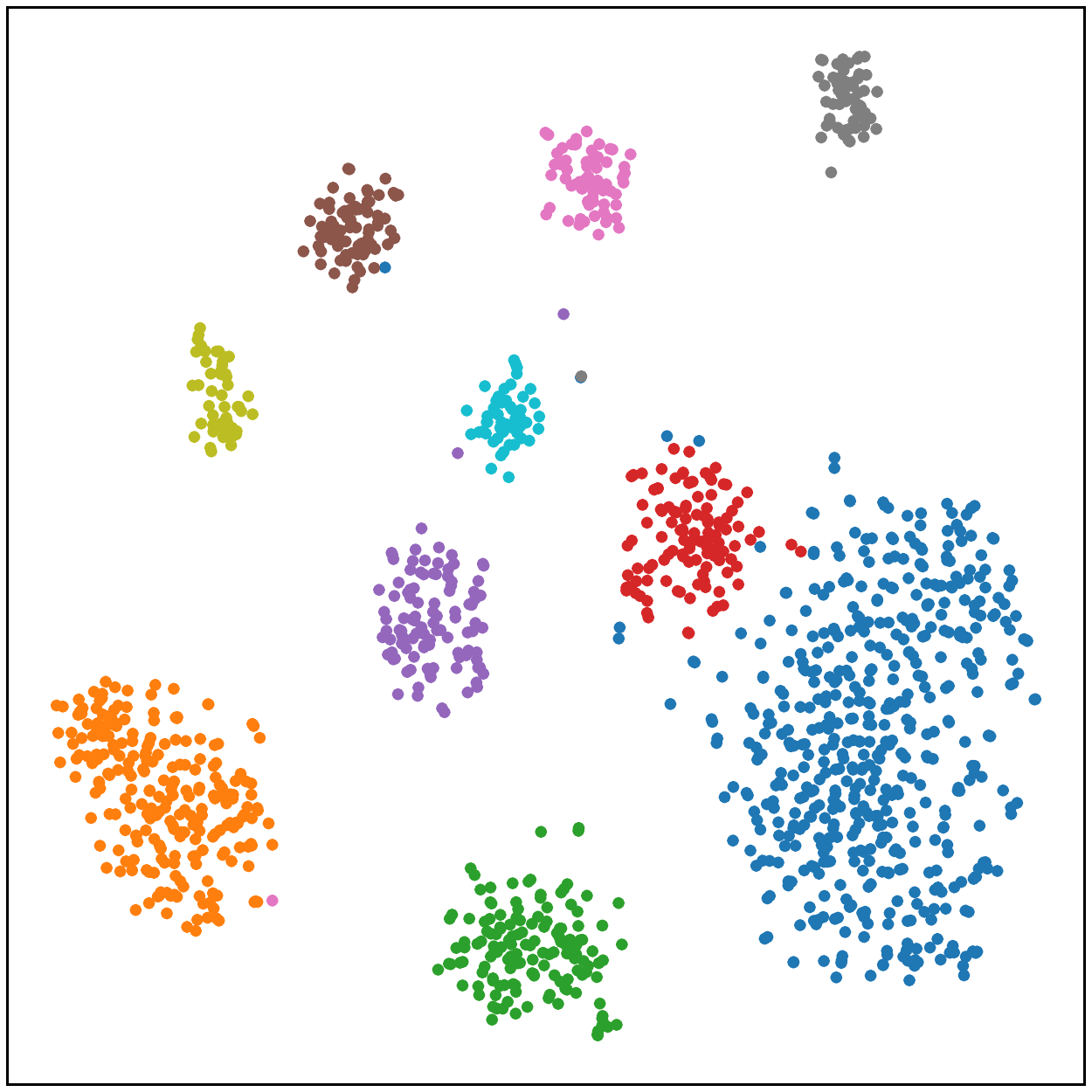}
  \includegraphics[width=1\textwidth]{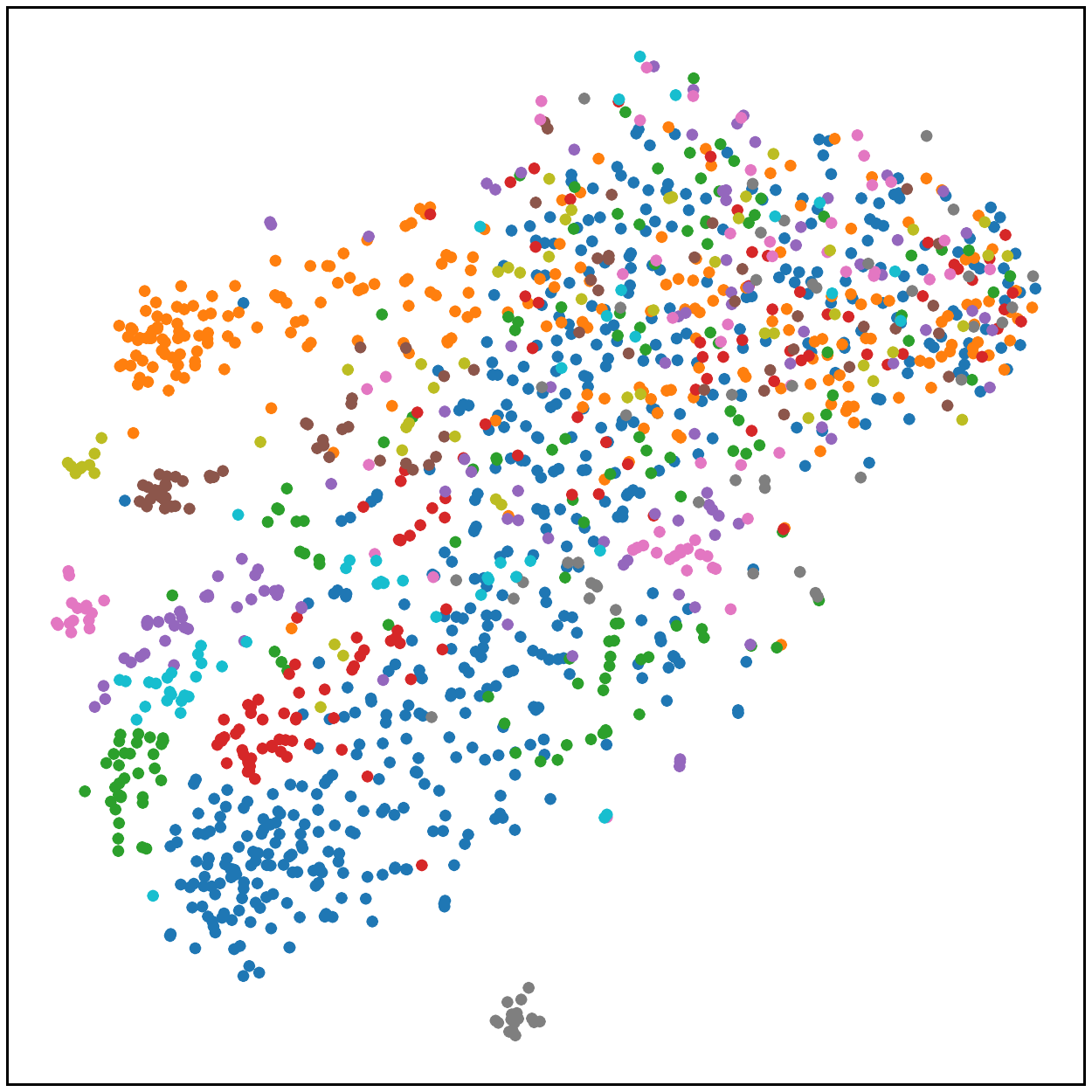}\vspace{0.3mm}
  \text{\footnotesize (g) C-SRIP}
 \end{minipage}\vspace{2.2mm}
 \begin{minipage}{0.015\textwidth}
  \begin{turn}{270}
  \footnotesize{\hspace{-5mm} MS \hspace{14mm} NMS}
  \end{turn}
 \end{minipage} 
 \caption{Visualization of ResNet-$101$ features extracted from hallucinated HR images using t-SNE~\cite{maaten2008visualizing}. Results are shown for the $10$ largest classes of  LFW. The plots show distributions for: (a) the original HR images, and (b-g) hallucinated HR face images images down-sampled using the matching (MS) or non-matching (NMS) degradation schemes. Best viewed in color and zoomed in.\vspace{-1.6mm}}
 \label{fig:tsne}
\end{figure*}

For the experiments with artificially down-sampled LFW data, we consider two different degradation schemes:
\begin{itemize}[leftmargin=*]\vspace{-0.5mm}
\item \textbf{A matching scheme (MS),} where each full-resolution LFW image is first filtered with a Gaussian kernel of $\sigma_b= 10$ and
the generated image is then decimated to the target size using bicubic interpolation. No noise is added. This scheme matches the training setup. \vspace{-1mm}
\item \textbf{A non-matching scheme (NMS)}, where $\sigma_b$ is selected randomly from a uniform distribution, i.e., $\mathcal{U}\left(0.5, 4\right)$, for each LFW image. 
After filtering and down-sampling, images are corrupted through additive Gaussian noise with standard deviation $\sigma_n$, drawn randomly from $\mathcal{U}(0, 20)$. This ensures a mismatch between the applied degradation function and the one used during training. Furthermore, it results in a different degradation for every test image.\vspace{-0.5mm}
\end{itemize}
The two schemes generate $24\times 24$ LR data of size and different characteristics as shown in Fig.~\ref{fig:face_samples_scface_lfw}. The generated images are then fed to the FH models for up-sampling and the HR results are used as inputs for ResNet-$101$.

For the experiments with the SCFace data, we use a subset of $650$ images captured by the five surveillance cameras at the largest of all recorded distances. After removing the interlaced rows from the images as well as a corresponding number of columns to ensure a correct aspect-ratio, we end up with images, where the facial area covers an image region close in size to the $24\times 24$ pixels expected by the FH models - a distribution for the SCFace face widths/heights is shown on the right of Fig.~\ref{fig:face_samples_scface_lfw}. We rescale  all images to the correct input size (using bicubic interpolation) and then feed the hallucination results produced by the FH models to ResNet-$101$ for descriptor computation.   

\textbf{Experiments on data separability.} Using the experimental setup described above, we explore whether data separability is improved when facial details are hallucinated 
how the separability is affected by the mismatch in the degradation function. To this end, we visualize the distribution of ResNet-$101$ feature descriptors extracted from hallucinated HR images of the $10$ largest LFW classes (i.e., the $10$ subjects with the highest number of images) using t-SNE~\cite{maaten2008visualizing} in Fig.~\ref{fig:tsne}. In order to quantitatively evaluate the separability of the presented distributions, we also compute a separability measure in the form of the Kullback-Leibler (KL) divergence between the distribution of a given class and joint distribution of all remaining classes in the 2D t-SNE embedding space and report average values calculated over all $10$ considered LFW classes in Table~\ref{tab: KL divergence}. 

We observe that the for the original HR images (before down-sampling) the classes are well separated and show no overlap. After down-sampling with the matching scheme (MS) and subsequent up-sampling (top row in Fig.~\ref{fig:tsne}), we see considerable overlap in the class distributions for bicubic interpolation. The FH models, on the other hand, improve the data separability over the interpolation-based baseline and result in significantly higher KL-divergence scores. C-SRIP performs particularly well and generates compact class clusters with very little overlap. 
\begin{table}[!tb]
\renewcommand{\arraystretch}{1.05}
\caption{Average KL divergence for the $10$ largest LFW classes with the MS and NMS degradation schemes estimated in the 2D space generated by t-SNE. Arrows indicate an increase or decrease in value compared to the baseline bicubic interpolation method.\vspace{1mm}} 
\label{tab: KL divergence}
\centering
\footnotesize
\begin{tabular}{ l  cccc}
\hline
\multirow{2}{*}{Approach}  & & \multicolumn{3}{c}{LFW}   \\\cline{3-5}
 && MS & NMS & Change \\ \hline
Bicubic (baseline) & &$0.5389$ &  $0.2135$ & $-0.3254$   \\
URDGN   & &$0.5561$ $\blu{\uparrow}$ & $0.2143$ $\blu{\uparrow}$ & $-0.3418$  \\
LapSRN   && $0.6346$ $\blu{\uparrow}$ & $0.2087$ $\red{\downarrow}$ & $-0.4259$ \\
CARN   & &$0.6851$ $\blu{\uparrow}$ & $0.1957$ $\red{\downarrow}$ & $-0.4894$ \\ 
SRResNet  & & $0.7148$ $\blu{\uparrow}$ & $0.1962$ $\red{\downarrow}$ & $-0.5222$  \\ 
C-SRIP   && $0.7676$ $\blu{\uparrow}$ & $0.1972$ $\red{\downarrow}$ & $-0.5704$  \\ \hline
\end{tabular}\vspace{-3mm}
\end{table} 

With the non-matching scheme (NMS) all mo\-dels perform noticeably worse, as shown in the bottom row of Fig.~\ref{fig:tsne}. Similarly as with the reconstruction experiments, we again see a drop in performance for bicubic interpolation, which is a learning-free  approach and was hence not trained for specific image characteristics. This suggests that ensuring good data separation is a harder task for LR images generated by NMS and that the drop in the KL divergence is not only a result of mismatched degradation functions. However, if we take the performance drop of the interpolation approach as our baseline, we observe that the FH models are much more sensitive to the characteristics of the LR data. The KL divergence of all models drops to a comparable value around $0.2$ and for the majority (except for URDGN) even falls slightly behind bicubic interpolation.

To further analyze the separability of the ResNet-$101$ descriptors of the hallucinated images, we report values for another non-parametric separability measure. i.e., Thornton's Geometric Separability Index (GSI), however, this time for the entire LFW and SCFace datasets and all FH models in Table~\ref{tab: GSI}. The index is defined as the fraction of data instances of a given dataset, $\mathcal{S}$, that has the same class-labels as their nearest neighbors, i.e.~\cite{thornton1998separability}: 
$GSI = \frac{1}{n}\sum_{i=1}^nf(\mathbf{z}_i,\mathbf{z}'_i)$,
where $n$ stands for the cardinality of $\mathcal{S}$ and $f$ is an indicator function that returns $1$ if the $i$-th ResNet-$101$ descriptor $\mathbf{z}_i$ and it's nearest neighbor $\mathbf{z}'_i$ share the same label and $0$ otherwise. GSI is bounded between $0$ and $1$, where a higher value indicates better separability. We use the cosine similarity to determine nearest neighbors.
\begin{table}[!tb]
\renewcommand{\arraystretch}{1.05}
\caption{GSI values achieved by the FH models in the ResNet-$101$ feature space. Note the decrease in the data separability due to mismatched degradation functions. Arrows indicate an increase or decrease in value compared to the baseline bicubic interpolation.\vspace{1mm}} 
\label{tab: GSI}
\centering
\footnotesize
\begin{tabular}{ l  ccccc}
\hline
\multicolumn{1}{l}{\multirow{2}{*}{Approach}}  & \multicolumn{3}{c}{LFW}  &\multirow{2}{*}{SCFace} \\\cline{2-4}
  & MS & NMS & Change & \\ \hline
  Bicubic (baseline) & $0.6283$ &  $0.5032$ &{$-19.9\%$}  & $0.5963$\\
 URDGN   & $0.6481$ $\blu{\uparrow}$ & $0.4866$ $\red{\downarrow}$ & {$-24,9\%$}   & $0.5346$\\
 LapSRN   & $0.6657$ $\blu{\uparrow}$ & $0.4906$ $\red{\downarrow}$& {$-26.3\%$}   & $0.6218$\\
 CARN   & $0.7130$ $\blu{\uparrow}$ & $0.4858$ $\red{\downarrow}$& {$-31.8\%$}  & $0.5691$\\ 
 SRResNet   & $0.7084$ $\blu{\uparrow}$ & $0.4927$ $\red{\downarrow}$& {$-30.4\%$}   & $0.5840$\\ 
 C-SRIP   & $0.7104$ $\blu{\uparrow}$ & $0.4893$ $\red{\downarrow}$& {$-31.1\%$}   & $0.5712$\\ \hline
\end{tabular}\vspace{-3mm}
\end{table} 

The results in Table~\ref{tab: GSI} again show that the data separability is improved with all FH models compared to the baseline with the MS scheme on LFW. With the NMS scheme all models perform worse than the baseline and also exhibit a larger drop in separability than simple bicubic interpolation. On SCFace we see a similar picture. Only LapSRN results in better separability than the interpolation-based baseline, while all other FH models decrease separability. 
These results again point to the importance of suitable training data, as FH models do not generalize well to unseen image characteristics and perform different than expected when applied on real-world imagery. 

\textbf{Recognition experiments.} In our last series of experiments we look at the recognition performance ensured by the FH models and extracted ResNet-$101$ descriptors on LFW and SCFace. For LFW we follow the so-called ``unrestricted outside data'' protocol and use the $6000$ pre-defined  image pairs in verification experiments. 
We keep one of the images in each pair unchanged (at the original resolution), and down-sample the second one using either the MS or NMS scheme. The LR images are then upscaled with the FH models and used to extract ResNet-$101$ descriptors. Matching is done with the cosine similarity. We report verification accuracy for the $10$ predefined experimental folds. For SCFace we perform a series of identification experiments, where we try to recognize subjects in the upscaled HR probe images based on the HR gallery data. 
\begin{figure}[!tb]
\begin{minipage}{0.49\columnwidth}
  \centering
  \includegraphics[width=1\textwidth]{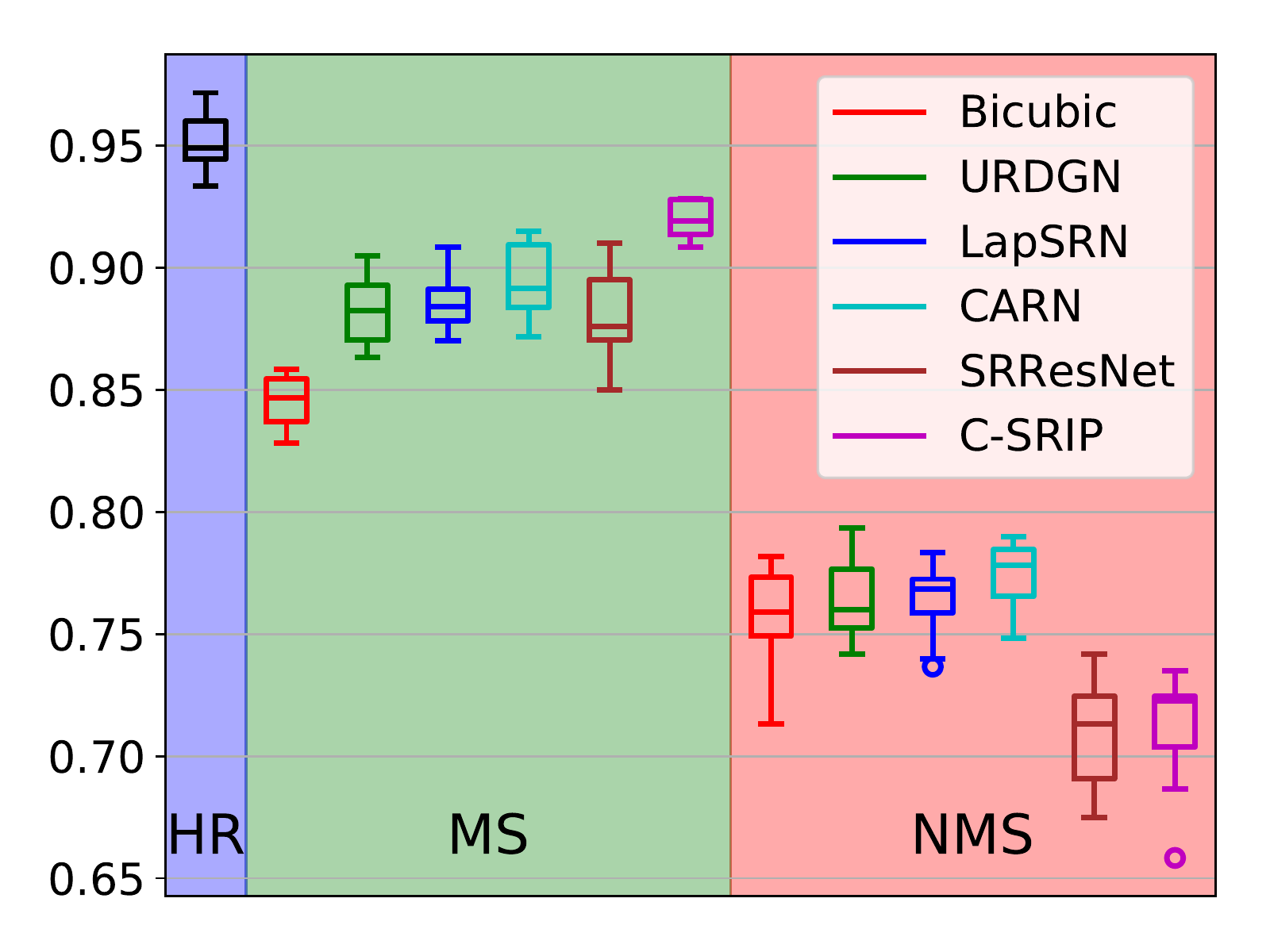}
\end{minipage}
\hfill
\begin{minipage}{0.49\columnwidth}
  \centering
  \vspace{1.5mm}
  \includegraphics[width=1\textwidth]{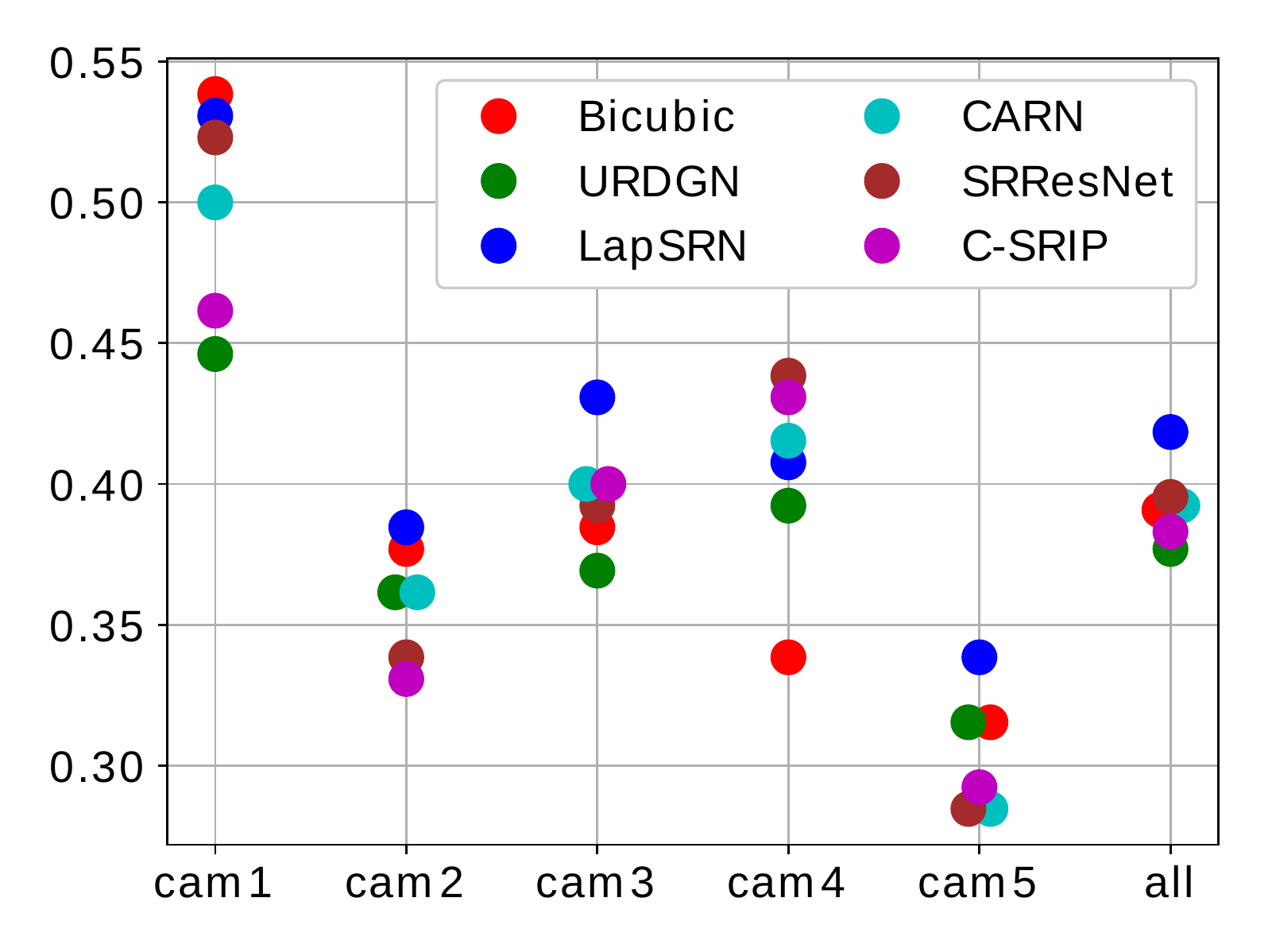}
  \end{minipage}\vspace{1mm}
  \caption{Recognition results on LFW (left) and SCFace (right). With a matching degradation function all models improve upon interpolation. The results are less predictable with image characteristics not seen during training. Best viewed in color.\vspace{-2mm}}
  \label{fig: LFW_SCFace_results}
\end{figure}

Fig.~\ref{fig: LFW_SCFace_results} shows that on HR LFW images the ResNet-$101$ model achieves a median verification accuracy of $95.1\%$. When the image size is reduced to $24\times24$ pixels with the MS scheme and the LR images are upscaled with bicubic interpolation, the performance drops to $84.5\%$. The FH models improve on this and achieve significantly better results. The highest median accuracy of $91.8\%$ comes from C-SRIP, which is the top performer in this setting. With the NMS scheme the drop in performance is larger for all methods  compared to the HR data. URDGN, LapSRN and CARN are only able to match the performance achieved by bicubic interpolation, while SRResNet and C-SRIP degrade results.

Results for SCFace  are shown  separately for each of the five cameras and in the form of the overall mean identification accuracy (i.e., rank-$1$) in Fig.~\ref{fig: LFW_SCFace_results}. We see that none of the FH models outperforms the bicubic baseline on all cameras. Overall, LapSRN offers a slight improvement over bicubic interpolation considering the average identification accuracy, but the performance gain is modest and in the range of $3\%$. The ranking of the models is also not consistent across different cameras, which generate LR data with very different characteristics. Observe, for example, C-SRIP, which performs worst with images from camera $2$, but is one of the top performers on camera $4$, where it gains around $10\%$ in performance over bicubic interpolation. 
These results show that without suitable mechanisms that are able to compensate for the bias introduced into FH model by the training data, hallucination results with real-world images are unpredictable and findings made with artificially down-sampled images cannot simply be extrapolated to real-world data.  

\section{Conclusion, discussion and outlook}

We have studied the impact of dataset bias on the problem of face hallucination and analyzed five recent CNN-based FH models on artificially degraded as well as real-world LR images. Below is summary of the main findings:
\begin{itemize}[leftmargin=*]\vspace{-0.5mm}
    \item \textbf{Reconstruction and robustness:} FH models achieve better reconstruction performance than the learning-free interpolation baseline on LR images matching the training data in terms of characteristics. However, their superiority fades away quickly 
    as the LR image characteristics diverge from the training setting. 
    The rather sudden drop in reconstruction quality points to an accuracy-robustness trade-off with FH models not present with learning-free approaches, as also observed for other CNN-based models by recent studies~\cite{stutz2018disentangling,su2018robustness}.\vspace{-0.5mm}
    \item \textbf{Separability and recognition:} We observe statistically significant improvements in data separability and face recognition performance, when LR image sre pre-processed with FH models (as opposed to interpolated), but \textit{only} for LR images degraded with the same approach as used during training. For mismatched image characteristics (with real-world data) we found no significant improvements in separability or recognition performance for any of the FH models, which in most cases fall behind simple interpolation.
\end{itemize}
Overall, our results suggest that despite recent progress, FH models are still very sensitive to the characteristics of the LR input data. We found limited empirical proof of their usefulness for higher-level vision tasks (e.g., recognition) beyond improvements in perceptually quality -- which might be important for representation-oriented problems, such as alignment or detection. Our analysis shows that we, as a community, need to move away from the standard evaluation methodology involving artificially degraded LR images and focus on more challenging real-world data when developing FH models for specific vision problems.

A common way to mitigate the effects of dataset bias in CNN-based models from the literature are domain adaption (DA) techniques or ensemble approaches \cite{csurka2017domain}. These have not been explored extensively for the problem of face hallucination yet (see~\cite{bulat2018learn} for initial attempts), but seem like an good starting point to improve the generalization abilities of FH models and make them applicable to real-world data.

{\small
\bibliographystyle{ieee}
\bibliography{egbib}
}

\clearpage

\appendix{}

\section{Appendix}

In this section we show some additional results and present additional findings made during our analysis.


\subsection{Reconstruction vs. recognition}

The literature on face hallucination typically focuses on designing FH models that ensure visually convincing super-resolution results and then presents recognition experiments with artificially down-sampled images to demonstrate how the models aid recognition. The main assumption here is that better reconstruction capabilities (in terms of average PSNR and SSIM scores) translate into better recognition performance. While this may be true for human perception, machine learning models do not necessarily behave in the same way, especially if the bias introduced into the models by the training data is taken into account.     

To analyze the relationship between reconstruction capabilities and the recognition performance ensured by the FH models, we present a number of results in Tables~\ref{tab: PSNR_SSIM_ms_nms},~\ref{tab: recons and recogn.} and~\ref{tab: Recognition rates ranking}. Table~\ref{tab: PSNR_SSIM_ms_nms} shows the average PSNR and SSIM scores for the matching (MS) and non-matching degradation (NMS) schemes achieved by the models on the LFW dataset. Table~\ref{tab: recons and recogn.} presents the recognition accuracy for both degradation schemes on LFW and separately for all five cameras on SCFace. Here, results are reported in terms of the average verification accuracy computed over $10$ experimental folds for LFW and as the rank-$1$ recognition rate for SCFace. Tables~\ref{tab: Recognition rates ranking} summarizes the results from Tables~\ref{tab: PSNR_SSIM_ms_nms} and ~\ref{tab: recons and recogn.} in terms of relative ranking for the given task.

From the presented results we see that the reconstruction quality with matching LR image characteristics is not a good indicator of the reconstruction quality with mismatched characteristics nor of the recognition performance ensured by the FH models. Models that performed well in one aspect do not necessarily generalize well to other tasks and image characteristics. C-SRIP, for example, achieves the highest PSNR and SSIM scores with the MS scheme on LFW and also leads to the best recognition performance in this setting, but performs worst in terms of reconstruction and recognition with the NMS scheme. Moreover, it also performs poorly on the SCFace data. LapSRN, on the other hand, is among the bottom three performers with the MS scheme on LFW in terms of reconstruction quality, but does better in the reconstruction experiments with the NMS scheme - the ranking here should be interpreted with reservation, as all tested models achieve very similar average PSNR scores. In terms of recognition performance, LapSRN still ensures only average results with the MS scheme, but does somewhat better with the NMS scheme. However, in recogition experiments on real-world SCFace data, LapSRN is overall the top performer, ensuring slight (statistically non-significant) improvements over the interpolation baseline and doing relatively well with LR images of all five cameras. If we look at the results for bicubic interpolation, we see that with matching LR image characteristics this baseline exhibits the weakest reconstruction capabilities, but is more competitive with non-matching data characteristics. In terms of recognition performance, it comes in last with the MS scheme, third (out of six) with the NMS scheme, and is very competitive on  SCFace . 

\begin{figure*}[!tb]
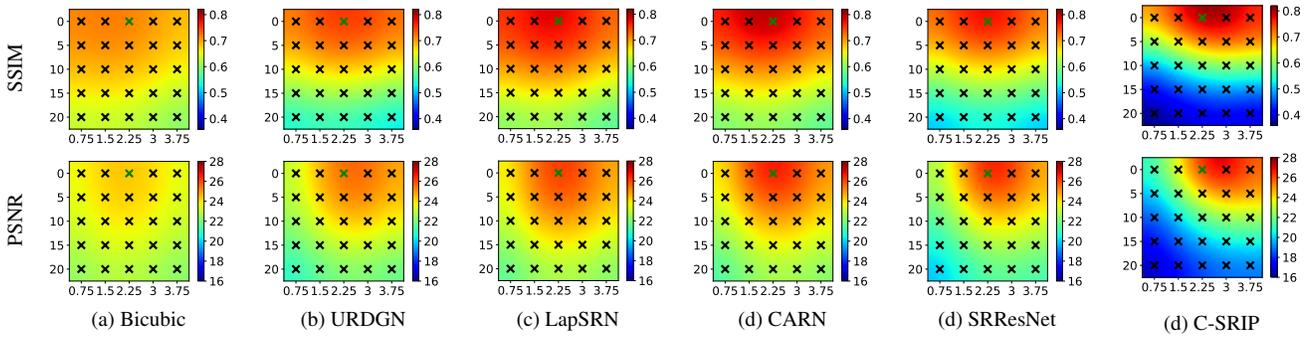

 \begin{minipage}{0.02\textwidth}
  \centering
  \begin{turn}{90}
\footnotesize{\hspace{4mm} PSNR \hspace{12mm} SSIM}
\end{turn}
 \end{minipage}
 \begin{minipage}{0.158\textwidth}
  \centering
  \includegraphics[width=1\textwidth,trim=1mm 0mm 0mm 0mm, clip]{reconstruction_grid_bicubic.pdf}\vspace{-2mm}
  \text{\footnotesize (a) Bicubic} 
 \end{minipage}
 \hfill
  \begin{minipage}{0.158\textwidth}
  \centering
  \includegraphics[width=1\textwidth,trim=1mm 0mm 0mm 0mm, clip]{reconstruction_grid_URDGN.pdf}\vspace{-2mm}
  \text{\footnotesize (b) URDGN} 
 \end{minipage}
 \hfill
  \begin{minipage}{0.158\textwidth}
  \centering
  \includegraphics[width=1\textwidth,trim=1mm 0mm 0mm 0mm, clip]{reconstruction_grid_LapSRN.pdf}\vspace{-2mm}
  \text{\footnotesize (c) LapSRN} 
 \end{minipage}
 \hfill
  \begin{minipage}{0.158\textwidth}
  \centering
  \includegraphics[width=1\textwidth,trim=1mm 0mm 0mm 0mm, clip]{reconstruction_grid_CARN.pdf}\vspace{-2mm}
  \text{\footnotesize (d) CARN} 
 \end{minipage}
 \hfill
   \begin{minipage}{0.158\textwidth}
  \centering
  \includegraphics[width=1\textwidth,trim=1mm 0mm 0mm 0mm, clip]{reconstruction_grid_srresnet.pdf}\vspace{-2mm}
  \text{\footnotesize (d) SRResNet} 
 \end{minipage}
 \hfill
   \begin{minipage}{0.158\textwidth}
  \centering
  \includegraphics[width=1\textwidth,trim=1mm 0mm 0mm 0mm, clip]{reconstruction_grid_C-SRIP.pdf}\vspace{-1mm}
  \text{\footnotesize (d) C-SRIP} 
 \end{minipage}\vspace{2mm}
\caption{Image reconstruction capabilities with mismatching degradation functions due to different blur and noise levels. The heat maps show the average SSIM (top row) and PSNR (bottom row) values computed over artificially degraded LFW images.  The points marked in the heat maps correspond to the different levels of noise ($\sigma_n$, increases vertically) and blur  ($\sigma_b$, increases horizontally). The value of $\sigma_n$ and $\sigma_b$ that was used for training is marked green. Note that all FH models achieve good reconstructions only around values that match the training setup. Best viewed in color and zoomed in.}\vspace{0mm}
\label{fig:heat_maps_all}
\end{figure*}

\begin{figure*}[!b]
\begin{minipage}{0.32\textwidth}
  \centering
  \includegraphics[width=1\textwidth]{downsampling_model_comparison_lr.pdf}\vspace{0.1mm}
  \text{\footnotesize (a) LR inputs ($\sigma_n$ vs. $\sigma_b$)}
 \end{minipage}
 \hfill
 \begin{minipage}{0.32\textwidth}
  \centering
  \includegraphics[width=1\textwidth]{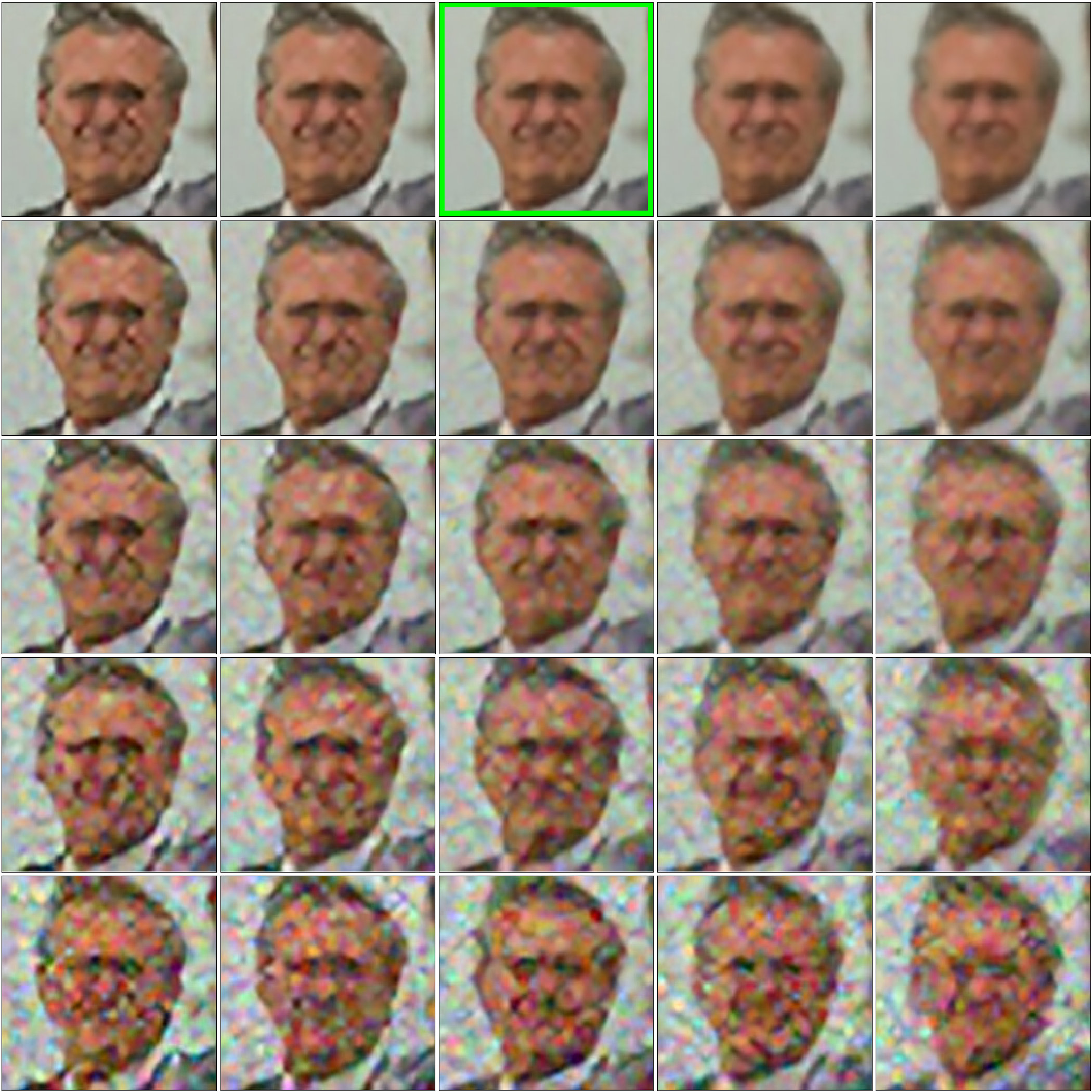}\vspace{0.1mm}
  \text{\footnotesize (b) URDGN ($\sigma_n$ vs. $\sigma_b$)}
 \end{minipage}
 \hfill
  \begin{minipage}{0.32\textwidth}
  \centering
  \includegraphics[width=1\textwidth]{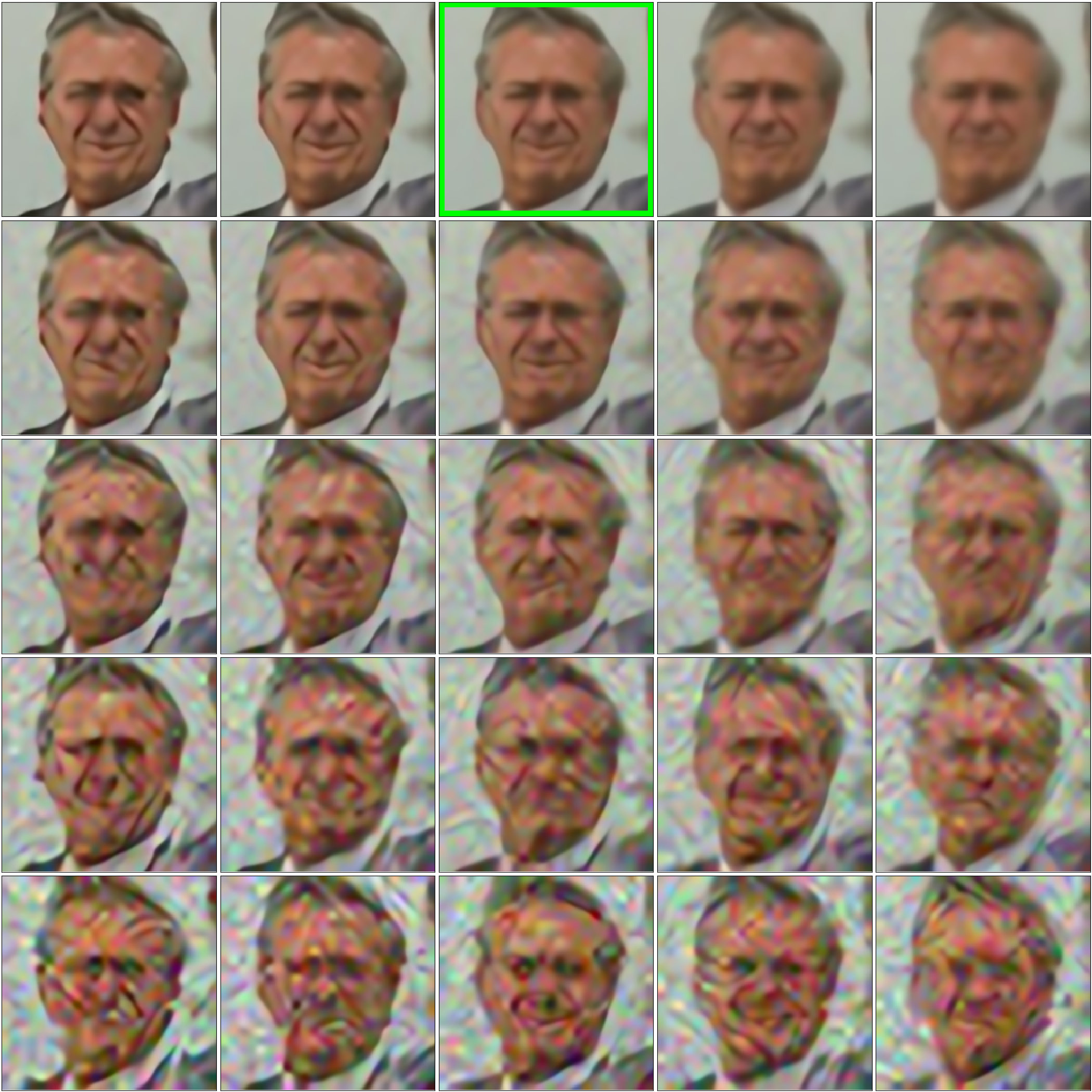}\vspace{0.1mm}
  \text{\footnotesize (c) LapSRN ($\sigma_n$ vs. $\sigma_b$)}
 \end{minipage}\vspace{2mm}
  \begin{minipage}{0.32\textwidth}
  \centering
  \includegraphics[width=1\textwidth]{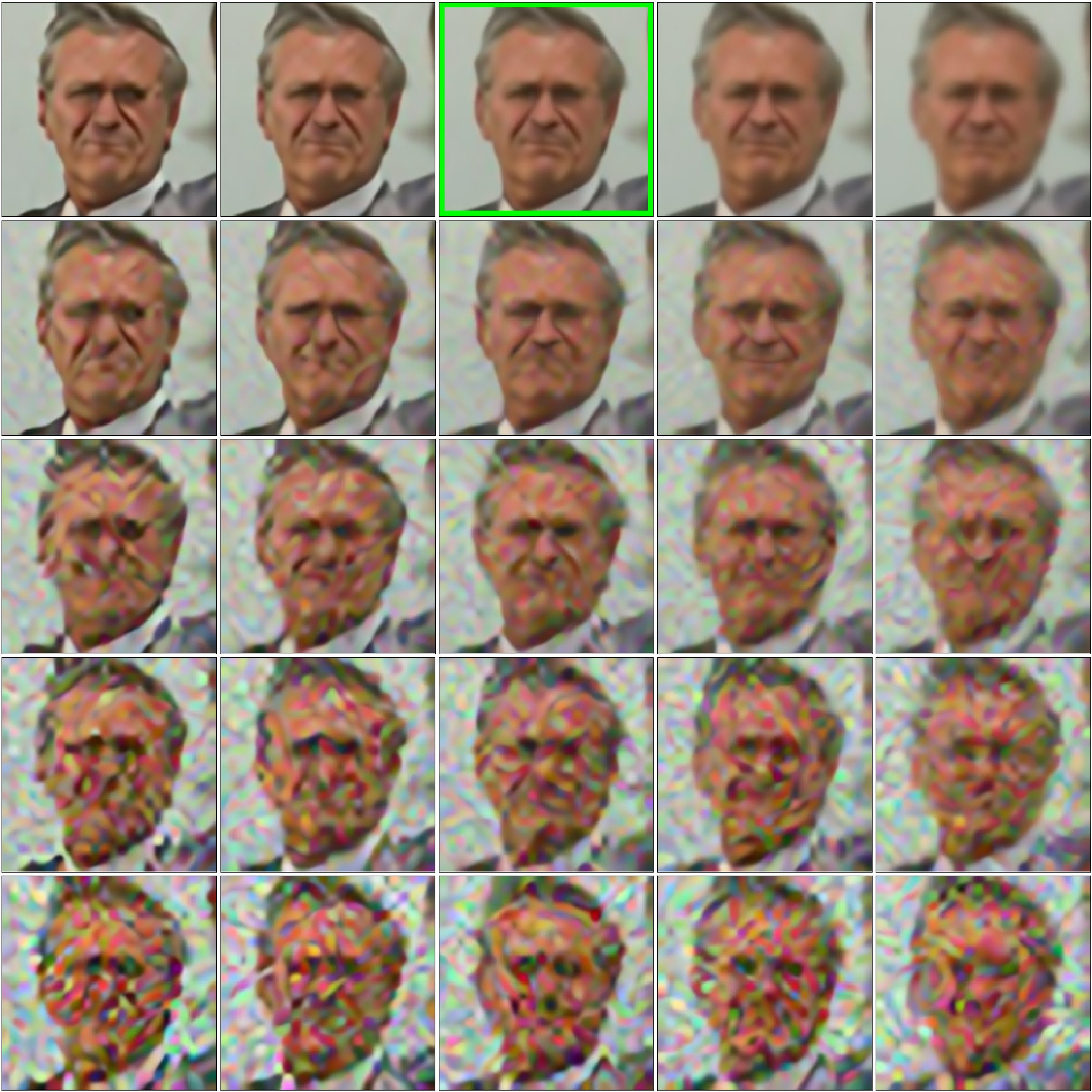}\vspace{0.1mm}
  \text{\footnotesize (d) SRResNet ($\sigma_n$ vs. $\sigma_b$)}
 \end{minipage}
 \hfill
 \begin{minipage}{0.32\textwidth}
  \centering
  \includegraphics[width=1\textwidth]{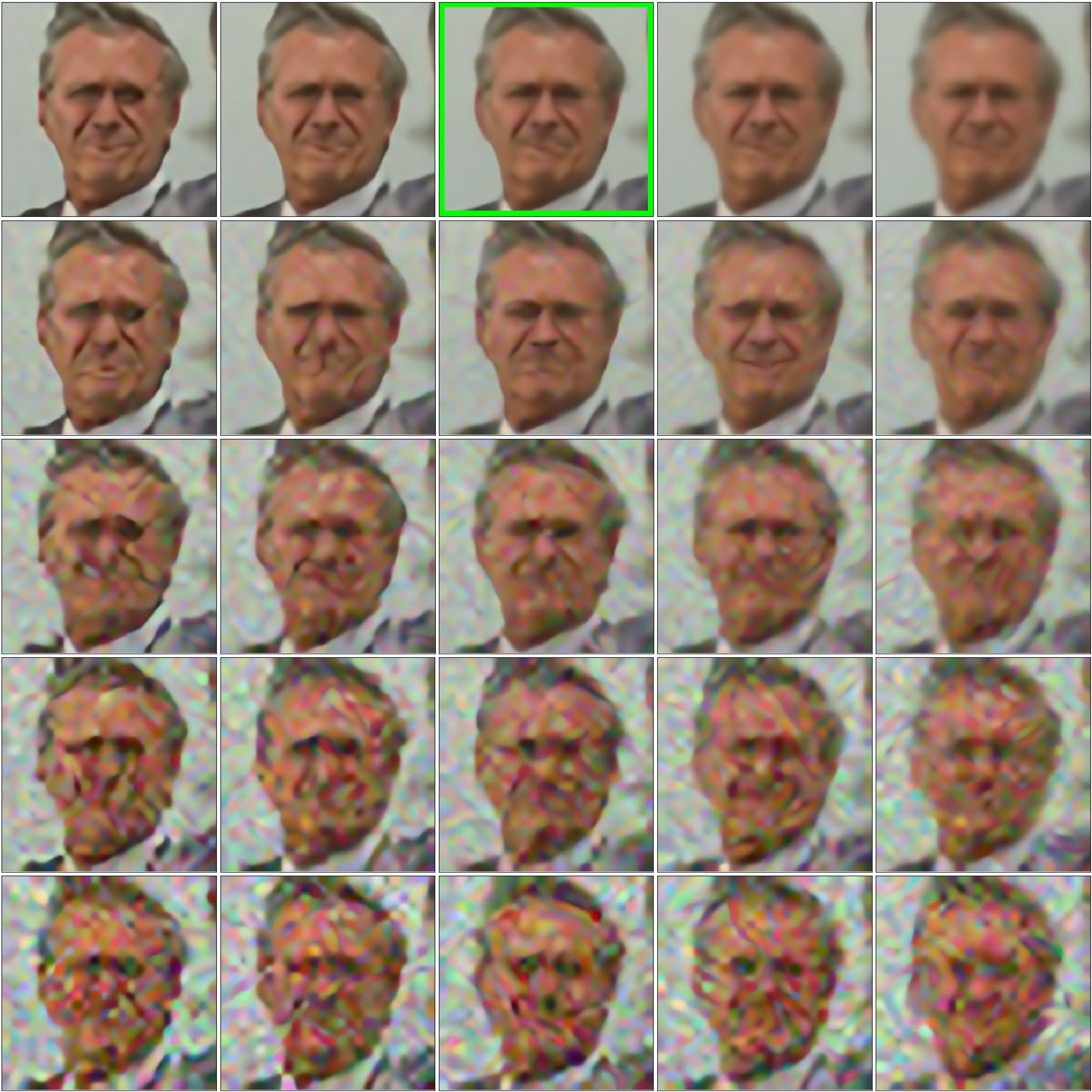}\vspace{0.1mm}
  \text{\footnotesize (e) CARN ($\sigma_n$ vs. $\sigma_b$)}
 \end{minipage}
 \hfill
  \begin{minipage}{0.32\textwidth}
  \centering
  \includegraphics[width=1\textwidth]{downsampling_model_comparison_C-SRIP.jpg}\vspace{0.1mm}
  \text{\footnotesize (f) C-SRIP ($\sigma_n$ vs. $\sigma_b$)}
 \end{minipage}\vspace{2mm}
\caption{Reconstruction capabilities of all tested FH models with mismatching degradation functions due to different blur and noise levels. The LR image block (with samples of size $24 \times 24$ pixels) in the top left corner illustrates the effect of increasing noise ($\sigma_n$, increases vertically) and blur ($\sigma_b$, increases horizontally) levels for a sample LFW image, the remaining image blocks show the $192 \times 192$ reconstructions generated by the tested models. Images marked green are generated with a degradation function matching the one used during training. Note that good HR reconstructions are achieved only with images degraded similarly as the training data. Best viewed zoomed in.}
\label{fig:SR_grids_noise_all}
\end{figure*}

Overall, we observe that the peak reconstruction performance achieved with matching image characteristics typically reported in the literature, does not correlate well with the recognition performance ensured by the FH models - when used as a preprocessing step for face recognition. Instead, we notice that the recognition performance seems to be related more to the robustness of the models and their ability to handle image characteristics not seen during training. If we look at the heat maps in Fig.~\ref{fig:heat_maps_all}, where SSIM and PSNR scores, computed over all LFW images, are presented for different noise and blur levels, we see that techniques that degrade the least across different settings, e.g., bicubic interpolation and LapSRN, in terms of reconstruction quality (even if their reconstruction performance is average), also result in competitive recognition accuracy with real-world images. Models, sensitive to image characteristics, such as C-SRIP, on the other hand, deteriorate quickly as the degradation function deviates from the training setup (see Fig.~\ref{fig:SR_grids_noise_all}), and perform relatively worse in the recognition task.  
\begin{table}[!tb]
\renewcommand{\arraystretch}{1.05}
\caption{Average SSIM and PSNR values achieved by the tested FH models with the matching (MS) and non-matching (NMS) degradation schemes on LFW.\vspace{1.5mm}} 
\label{tab: PSNR_SSIM_ms_nms}
\centering
\footnotesize
\begin{tabular}{ l  cccccc}
\hline
\multicolumn{1}{l}{\multirow{2}{*}{Approach}}  & \multicolumn{2}{c}{MS} & &\multicolumn{2}{c}{NMS} \\\cline{2-3} \cline{5-6}
  & PSNR & SSIM & & PSNR & SSIM\\ \hline
 Bicubic & $24.401$ & $0.7129$ && $23.459$ & $0.6745$\\
 URDGN   & $25.594$ & $0.7539$ && $23.434$ & $0.6566$\\
 LapSRN  & $26.417$ & $0.7792$ && $23.967$ & $0.6929$\\
 CARN    & $26.894$ & $0.7938$ && $23.934$ & $0.6923$\\ 
 SRResNet& $27.176$ & $0.8013$ && $23.589$ & $0.6760$\\ 
 C-SRIP  & $27.233$ & $0.8202$ && $23.138$ & $0.6145$\\ \hline
\end{tabular}
\end{table}

\begin{table}[!tb]
\renewcommand{\arraystretch}{1.05}
\caption{Average recognition accuracy achieved by the tested FH models and ResNet-$101$ with the matching (MS) and non-matching (NMS) degradation schemes on LFW and all cameras of SCFace at the largest $2.6$m distance.\vspace{1.5mm}} 
\label{tab: recons and recogn.}
\centering
\footnotesize
\resizebox{\columnwidth}{!}{%
\begin{tabular}{ l  ccccccccc}
\hline
\multicolumn{1}{l}{\multirow{2}{*}{Approach}}  & \multicolumn{2}{c}{LFW} & &\multicolumn{5}{c}{SCFace} \\\cline{2-3} \cline{5-9}
  & MS & NMS & & C1 & C2& C3& C4& C5\\ \hline
Bicubic  & $0.846$ & $0.759$ && $0.538$ & $0.377$ & $0.384$ & $0.338$ & $0.315$\\
URDGN    & $0.882$ & $0.759$ && $0.445$ & $0.361$ & $0.369$ & $0.392$ & $0.315$\\
LapSRN   & $0.884$ & $0.768$ && $0.531$ & $0.384$ & $0.431$ & $0.407$ & $0.338$\\
CARN     & $0.891$ & $0.778$ && $0.500$ & $0.361$ & $0.400$ & $0.415$ & $0.284$\\ 
SRResNet & $0.876$ & $0.713$ && $0.523$ & $0.338$ & $0.392$ & $0.438$ & $0.284$\\ 
C-SRIP   & $0.919$ & $0.722$ && $0.461$ & $0.330$ & $0.400$ & $0.431$ & $0.292$\\ \hline
\end{tabular}}
\end{table}

\begin{table*}[!tb]
\renewcommand{\arraystretch}{1.05}
\caption{Relative ranking of the FH models for the reconstruction (based on PSNR) and recognition tasks with matched and mismatched LR image characteristics. Note that good reconstruction performance does not necessarily translate into good recognition performance. \vspace{2mm}} 
\label{tab: Recognition rates ranking}
\centering
\footnotesize
\begin{tabular}{ l  cccccccccccc}
\hline
\multicolumn{1}{l}{\multirow{2}{*}{Approach}}  & \multicolumn{2}{c}{LFW - Reconstruction} & & \multicolumn{2}{c}{LFW - Recognition} & &\multicolumn{5}{c}{SCFace - Recognition} \\\cline{2-3} \cline{5-6} \cline{8-12}
         & MS  & NMS && MS  & NMS && C1  & C2  & C3  & C4  & C5\\ \hline
 Bicubic & 6th & 4th && 6th & 3rd && 1st & 2nd & 5th & 6th & 2nd\\
 URDGN   & 5th & 5th && 4th & 4th && 6th & 3rd & 6th & 5th & 3rd\\
 LapSRN  & 4th & 1st && 3rd & 2nd && 2nd & 1st & 1st & 4th & 1st\\
 CARN    & 3rd & 2nd && 2nd & 1st && 4th & 4th & 2nd & 3rd & 5th\\ 
 SRResNet& 2nd & 3rd && 5th & 5th && 3rd & 5th & 4th & 1st & 6th\\ 
 C-SRIP  & 1st & 6th && 1st & 6th && 5th & 6th & 3rd & 2nd & 4th\\ \hline
\end{tabular}
\end{table*}

While we do not explore these findings further, our results suggest that the common way of optimizing the peak reconstruction performance of FH models may not be the most optimal choice of approaching the hallucination problem if the target application is face recognition. In the recently observed accuracy-robustness trade-off of CNN models~\cite{stutz2018disentangling,su2018robustness}, the robustness aspect seems more important, even, when it comes at the expense of performance. This is an interesting observation and suggests that we need to rethink the standard methodologies used in the field of face hallucination, especially if the hallucination task is paired with a higher-level vision problem. 

\begin{figure*}[!t]
    \centering
    \begin{minipage}{0.02\textwidth}
        \begin{turn}{90}
            \scriptsize{\hspace{5mm}cam5\hspace{7mm} cam4\hspace{7mm} cam3\hspace{7mm} cam2\hspace{7mm} cam1}
        \end{turn}
    \end{minipage}
    \begin{minipage}{0.52\textwidth}
        \includegraphics[width=1\textwidth]{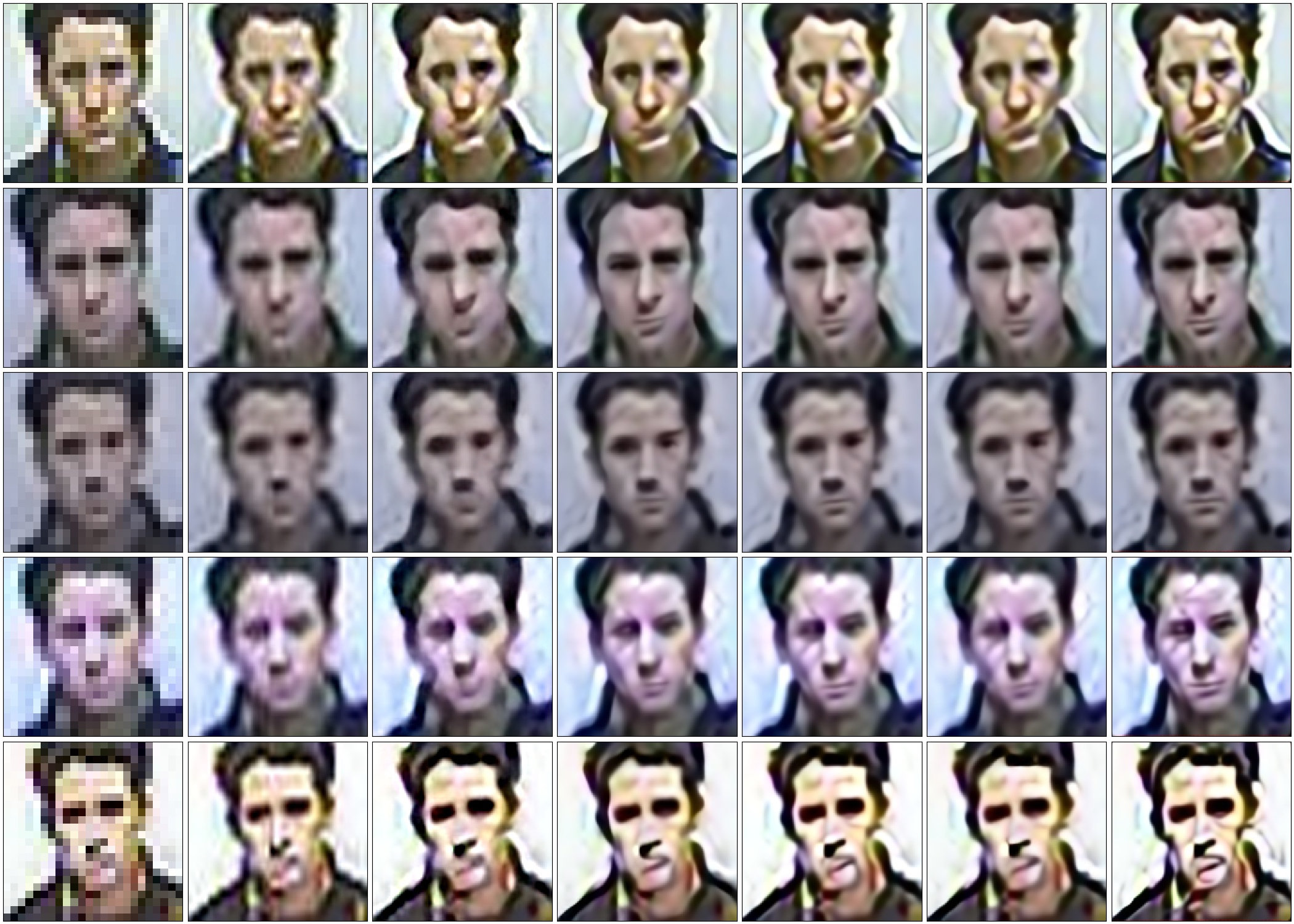}
        \scriptsize{\hspace*{5mm}LR\hspace{7mm} Bicubic \hspace{3.5mm} URDGN \hspace{3mm} LapSRN \hspace{4mm} CARN \hspace{3mm} SRResNet \hspace{3mm} C-SRIP}
    \end{minipage}
    \hspace{7mm}
    \begin{minipage}{0.24\textwidth}
        \vspace{0.2mm}
        \centering
        \includegraphics[width=1\textwidth]{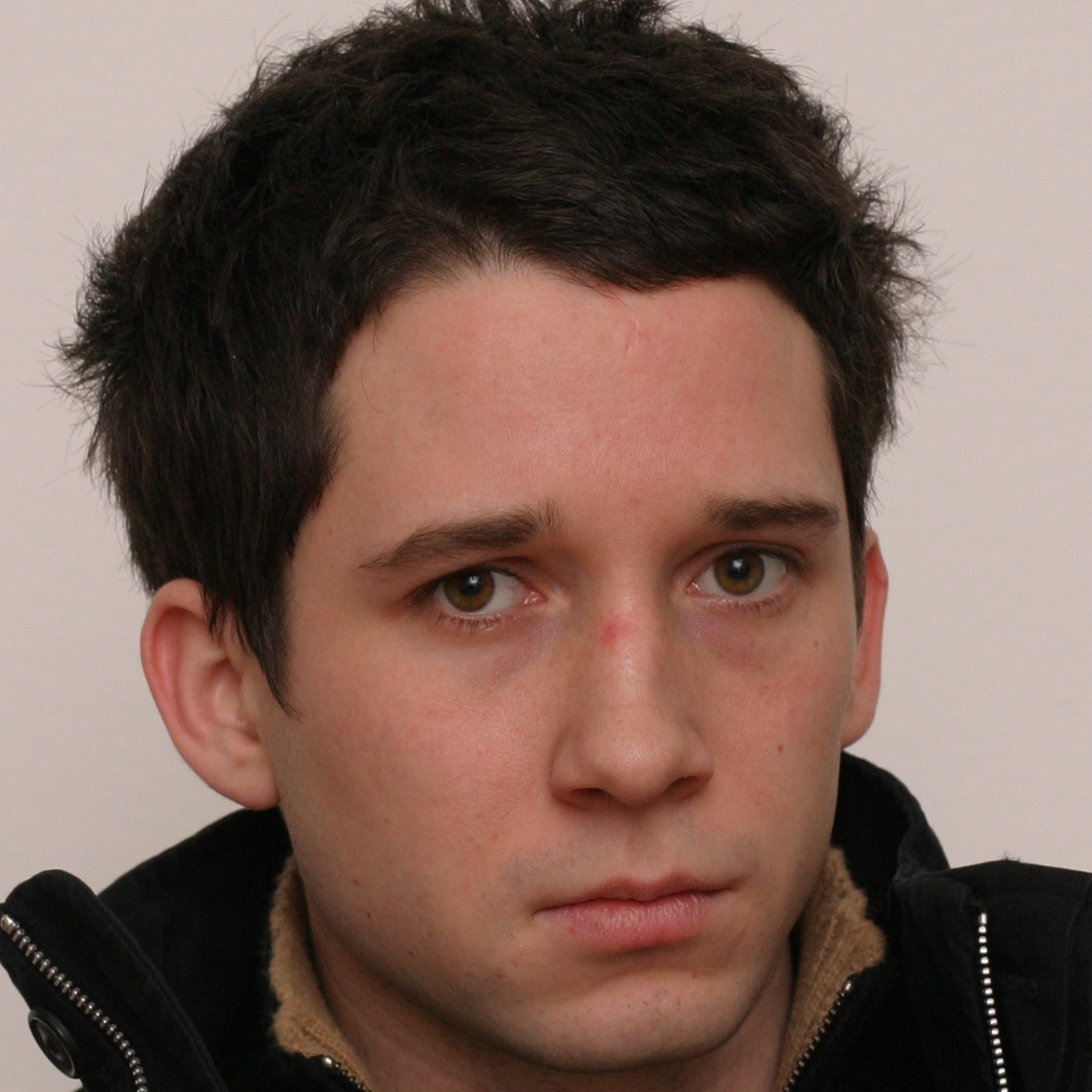}
        \scriptsize{Subject gallery image}
    \end{minipage}\\
    \vspace{2.5mm}
    \begin{minipage}{0.02\textwidth}
        \begin{turn}{90}
            \scriptsize{\hspace{5mm}cam5\hspace{7mm} cam4\hspace{7mm} cam3\hspace{7mm} cam2\hspace{7mm} cam1}
        \end{turn}
    \end{minipage}
    \begin{minipage}{0.52\textwidth}
        \includegraphics[width=1\textwidth]{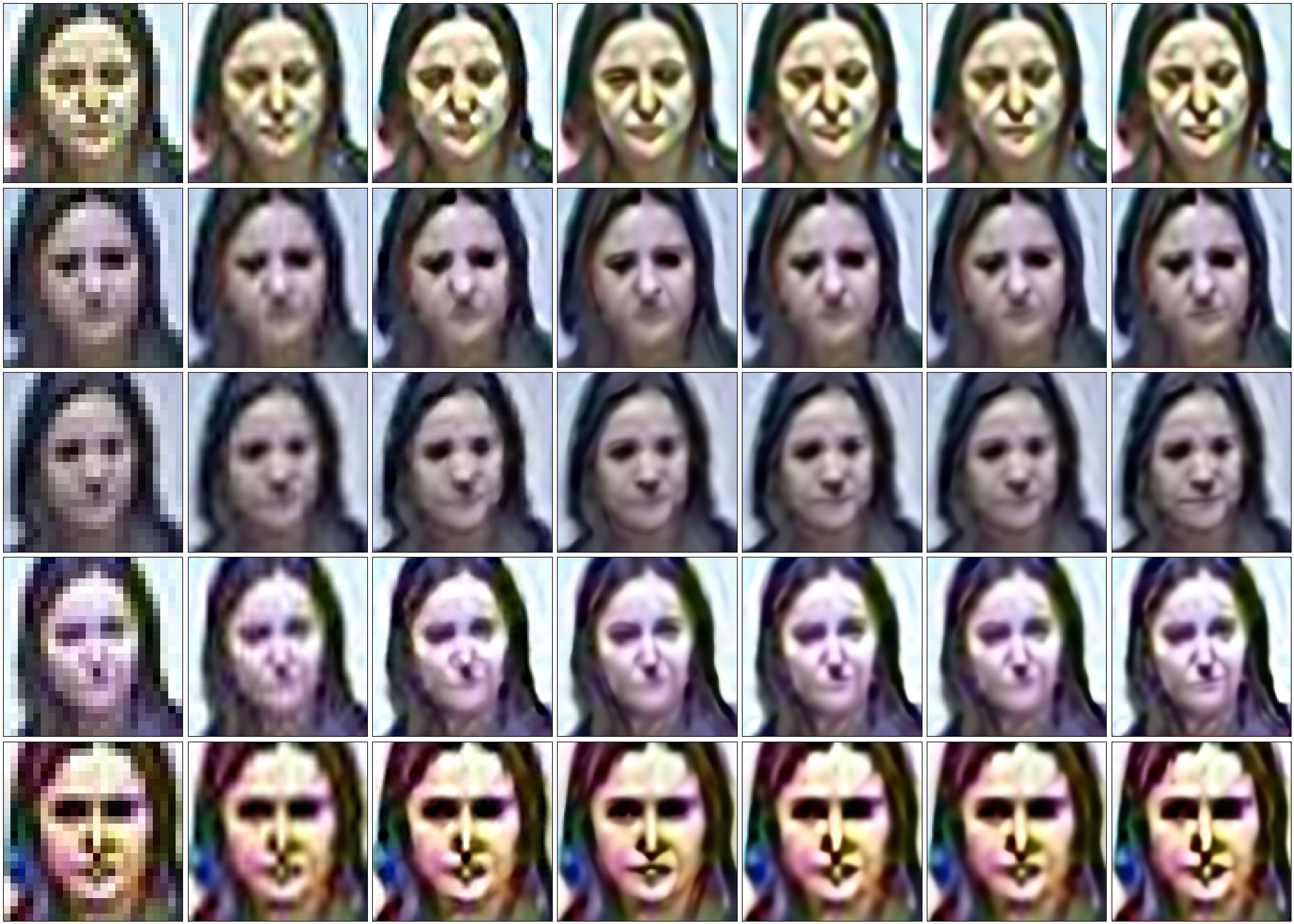}
        \scriptsize{\hspace*{5mm}LR\hspace{7mm} Bicubic \hspace{3.5mm} URDGN \hspace{3mm} LapSRN \hspace{4mm} CARN \hspace{3mm} SRResNet \hspace{3mm} C-SRIP}
    \end{minipage}
    \hspace{7mm}
    \begin{minipage}{0.24\textwidth}
        \vspace{0.2mm}
        \centering
        \includegraphics[width=1\textwidth]{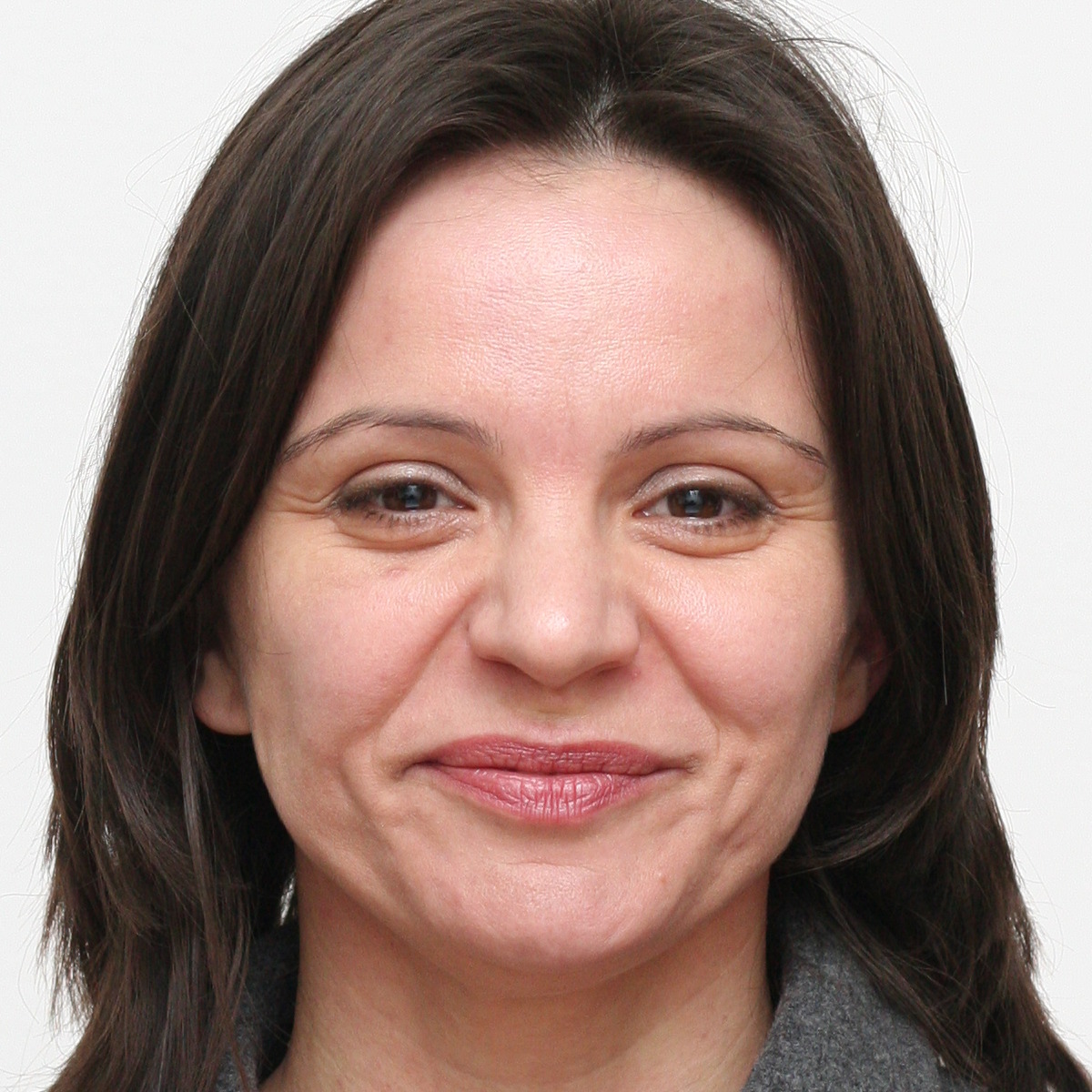}
        \scriptsize{Subject gallery image}
    \end{minipage}\\
    \vspace{2.5mm}
    \begin{minipage}{0.02\textwidth}
        \begin{turn}{90}
            \scriptsize{\hspace{5mm}cam5\hspace{7mm} cam4\hspace{7mm} cam3\hspace{7mm} cam2\hspace{7mm} cam1}
        \end{turn}
    \end{minipage}
    \begin{minipage}{0.52\textwidth}
        \includegraphics[width=1\textwidth]{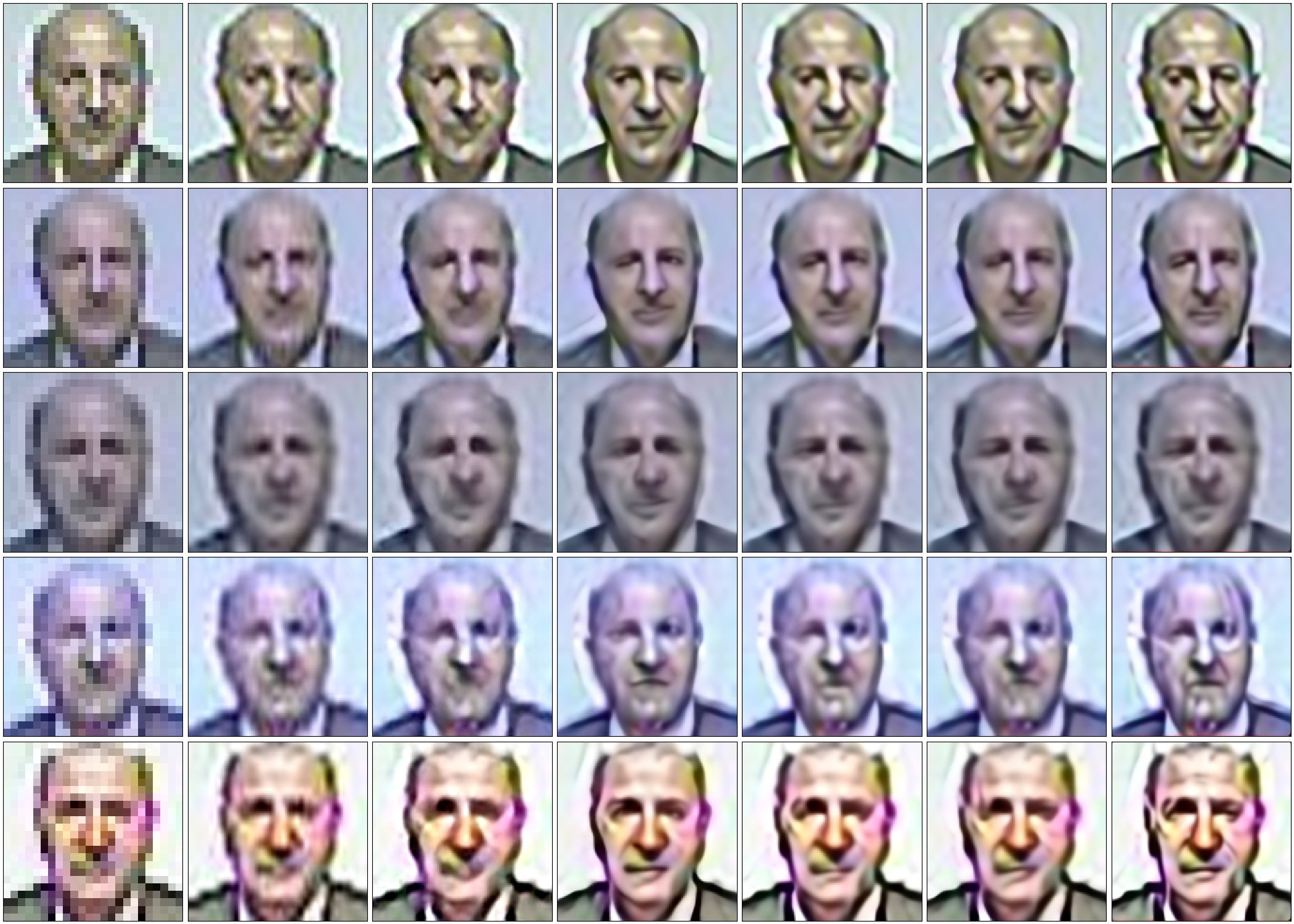}
        \scriptsize{\hspace*{5mm}LR\hspace{7mm} Bicubic \hspace{3.5mm} URDGN \hspace{3mm} LapSRN \hspace{4mm} CARN \hspace{3mm} SRResNet \hspace{3mm} C-SRIP}
    \end{minipage}
    \hspace{7mm}
    \begin{minipage}{0.24\textwidth}
        \vspace{0.2mm}
        \centering
        \includegraphics[width=1\textwidth]{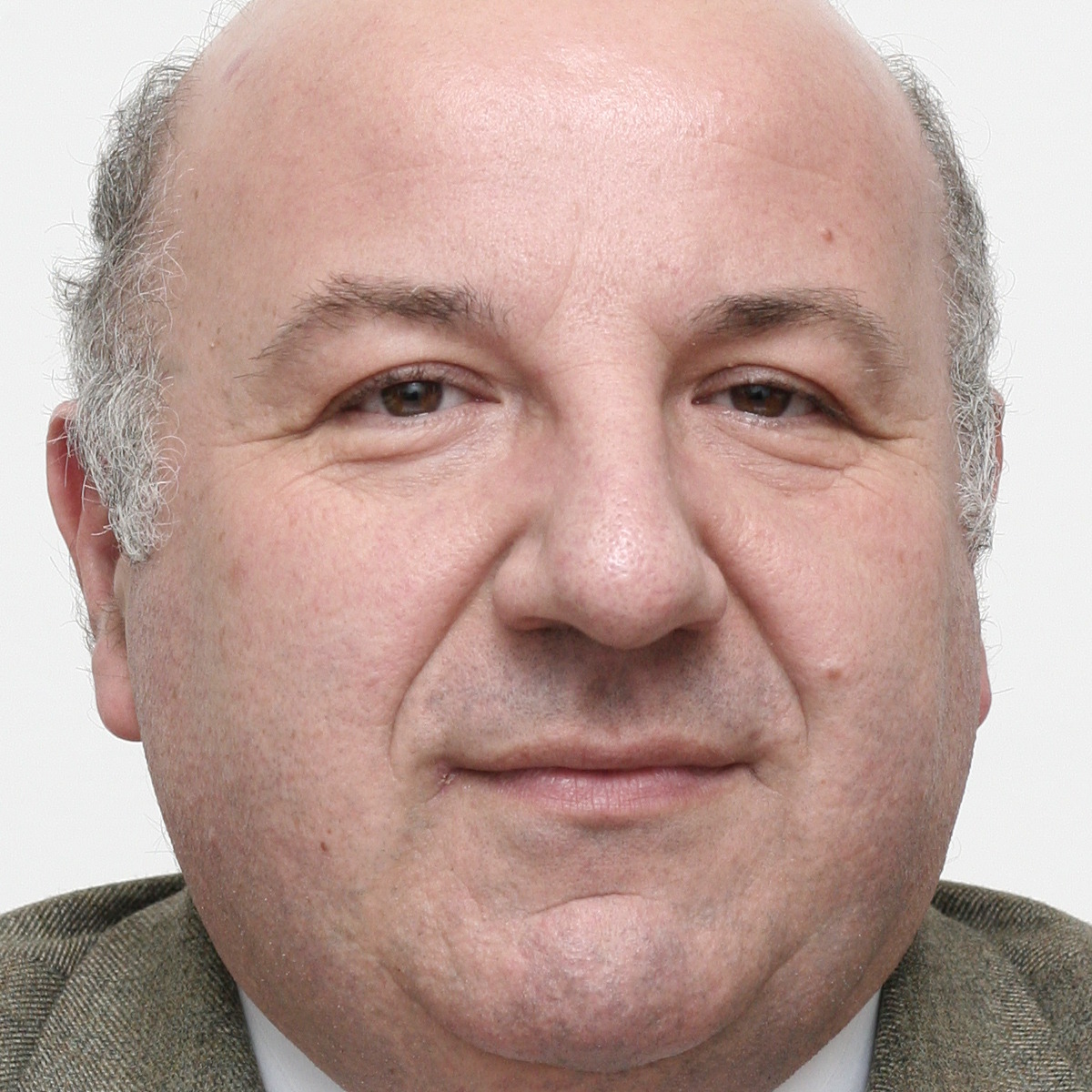}
        \scriptsize{Subject gallery image}
    \end{minipage}\\
    \vspace{2.5mm}
    \caption{Examples of  hallucinated HR faces from the SCFace~\cite{grgic2011scface} dataset. Results are presented for three subjects and separately for all five surveillance cameras. The FH models offer improvements in discernible facial details over the interpolation baseline, but introduce considerable distortions for some of the cameras (i.e., cameras $1$ and $5$). The figure is best viewed zoomed in.}
    \label{fig:scface_sr}
\end{figure*}

\subsection{Mismatch due to blur and noise}

In Fig.~\ref{fig:SR_grids_noise_all} we show visual examples of the reconstruction capabilities of all tested FH model, as opposed to the main part of the paper, where only results for the bicubic interpolation approach and the top-performer in terms of reconstruction quality with matching image characteristics, i.e., the C-SRIP model, were presented. We see a consistent behaviour with all models - they are able to generate convincing reconstructions with LR images matching the characteristics of the data used during training, but introduce considerable artifacts as soon as the degradation function starts to deviate from the training degradation function. Also, we observe that the visual quality of the reconstructions produced by C-SRIP, which produces the highest quality HR face images with a matching function, is most affected by the mismatch. The remaining models still deteriorate in performance, but visually, the results appear less distorted.

More objective results evaluating the effect of mismatched noise and blur levels over the entire LFW dataset are presented in the heat maps in Fig.~\ref{fig:heat_maps_all}. Here, we show results for PSNR and SSIM, while in the main part of the paper only heat maps for SSIM were presented. The point that corresponds to the training setting is again marked green in the figures. We see that PSNR behaves similar to SSIM. the FH models are able to outperform bicubic interpolation significantly around conditions similar to the ones seen during training, but are less robust than interpolation and degrade faster once the degradation function used to generate the LR data starts to deviate from the functions used during training.   

\subsection{Reconstructing real-world LR images}
In Fig.~\ref{fig:scface_sr}, we present outputs of the tested face hallucination models on the SCFace~\cite{grgic2011scface} dataset. As in our recognition experiment, we only consider the images captured from a $2.6$ meter distance (i.e., the distance $1$ series of the dataset), because these images most closely match the training input size of the face hallucination models. As this dataset contains real-world images from several different surveillance cameras, we don't have a high-resolution ground truth images available for the LR images, only a high-resolution gallery for every subject. We therefore compare the the outputs of the face hallucination models to these HR galleries qualitatively. 

Fig.~\ref{fig:scface_sr} shows that the characteristics of the LR images differ considerably from camera to camera and affect not only the perceived quality of the LR data, but also other aspects, such as color scheme, saturation, image contrast, etc. The FH models are able to improve upon the visual quality of the HR reconstructions compared to the interpolation baseline for certain cameras and less so for others. For example, we see considerably more facial details in the C-SRIP reconstructions on cameras $2$, $3$, and $4$ compared to the baseline. On cameras $1$ and $5$, however, the images still appear crisper and less blurred, but image artifacts are also present and contribute to the perception of low-quality HR reconstructions. Similar observations can also be made for other FH models, which follow the outlined trend and behave similarly to C-SRIP. Considering the three selected examples, LapSRN seems to strike a good balance between reconstruction quality and the amount of introduced image distortions - observe the HR reconstructions from cameras $1$ and $5$ for all three subjects.

\end{document}